\documentclass[journal]{IEEEtran}

\usepackage{url}
\usepackage{hyperref}
\usepackage{booktabs}
\usepackage{graphicx}
\usepackage{subfigure}
\usepackage{bm}
\usepackage{color}
\usepackage{amsmath}
\usepackage{amsfonts}
\usepackage{stmaryrd}
\usepackage{stfloats}
\usepackage{xspace}
\usepackage{multirow}
\usepackage{pifont}
\usepackage{pbox}
\usepackage{colortbl}
\usepackage{bbm}
\usepackage{bbding}
\usepackage{wrapfig}
\usepackage{balance}
\usepackage{makecell}

\usepackage{algorithm}
\usepackage{algpseudocode} % algorithmicx 的伪代码风格

\hypersetup{urlcolor=cyan}

\newcommand\ies{\textit{i.e.}}
\newcommand\ie{\textit{i.e.}}
\newcommand\egs{\textit{e.g.}}

\newcommand\figcaption{\def\@captype{figure}\caption}
\newcommand\tabcaption{\def\@captype{table}\caption}

\ifCLASSOPTIONcompsoc
\usepackage[nocompress]{cite}
\else
\usepackage{cite}
\fi

\begin{document}
%
% paper title
% Titles are generally capitalized except for words such as a, an, and, as,
% at, but, by, for, in, nor, of, on, or, the, to and up, which are usually
% not capitalized unless they are the first or last word of the title.
% Linebreaks \\ can be used within to get better formatting as desired.
% Do not put math or special symbols in the title.
% \title{Discriminating Noisy Pseudo Labels for Audible Video Event Parsing: An Evidence View}
\title{Truth in the Few: High-Value Data Selection for Efficient Multi-Modal Reasoning}
%
%
% author names and IEEE memberships
% note positions of commas and nonbreaking spaces ( ~ ) LaTeX will not break
% a structure at a ~ so this keeps an author's name from being broken across
% two lines.
% use \thanks{} to gain access to the first footnote area
% a separate \thanks must be used for each paragraph as LaTeX2e's \thanks
% was not built to handle multiple paragraphs
%

 \author{
    Shenshen Li, Xing Xu ~\IEEEmembership{Member,~IEEE}, Kaiyuan Deng, Lei Wang, Heng Tao Shen ~\IEEEmembership{Fellow,~IEEE}, Fumin Shen
%   \thanks{This work was supported by the National Natural Science Foundation of China under Grants (No. 62222203 and 62072080), the New Cornerstone Science Foundation through the XPLORER PRIZE and the Sichuan Science and Technology Program (No. 2023-XT00-00001-GX). (Corresponding author: Xing Xu.)}
%   \thanks{S. Li, W. Meng, Y. Yang, and F. Shen are with the Center for Future Media and School of Computer Science and Engineering, University of Electronic Science and Technology of China, Chengdu 611731, China (E-mail: lishenshen727@gmail.com; chen\_he1229@outlook.com; yang.yang@uestc.edu.cn; fumin.shen@gmail.com).}
%   \thanks{X. Xu and H. T. Shen are with the Center for Future Media, School of Computer Science and Engineering, University of Electronic Science and Technology of China, Chengdu 611731, China and also with with College of Electronic and Information Engineering, Tongji University, Shanghai 201804, China (e-mail: interxuxing@hotmail.com; shenhengtao@hotmail.com).}
	\thanks{This work was supported by the National Natural Science Foundation of China under Grants (No. 62476201), Chengdu Science and Technology Program (Grant No. 2023-XT00-00001-GX), Chengdu Science and Technology Bureau (Grant No. 2025ZDZX0008), the New Cornerstone Science Foundation through the XPLORER PRIZE (Corresponding author: Fumin Shen.)}
	\thanks{Shenshen Li, Kaiyuan Deng and Fumin Shen are with the School of Computer Science and Engineering, University of Electronic Science and Technology of China, Chengdu 611731, China. (E-mail: lishenshen727@gmail.com, mantou.cloud@gmail.com, fumin.shen@gmail.com).}
	\thanks{L. Wang is with MiroMind in Singapore 408621 (E-mail: lei.wang.2019@phdcs.smu.edu.sg).}
	\thanks{X. Xu and H. T. Shen are with the School of Computer Science and Technology, Tongji University, Shanghai 201804, China. (E-mail: interxuxing@hotmail.com, shenhengtao@hotmail.com).}
 }

% The paper headers
\markboth{Preprint. Under review.}%
{Shell \MakeLowercase{\textit{et al.}}: Bare Demo of IEEEtran.cls for IEEE Journals}

% make the title area
\maketitle

	\begin{abstract}
		While multi-modal large language models (MLLMs) have made significant progress in complex reasoning tasks via reinforcement learning, it is commonly believed that extensive training data is necessary for improving multi-modal reasoning ability, inevitably leading to data redundancy and substantial computational costs.
		% However, \textit{does scaling the data size inherently improve cross-modal reasoning in MLLMs?}
		However, \textit{can smaller high-value datasets match or outperform full corpora for multi-modal reasoning in MLLMs?}
		In this work, we challenge this assumption through a key observation: meaningful multi-modal reasoning is triggered by only a sparse subset of training samples, termed \textit{cognitive samples}, whereas the majority contribute marginally.
		% Building on this insight, we propose a novel data selection framework termed \textit{\textbf{R}easoning \textbf{A}ctivation \textbf{P}otential (RAP)} for more efficient cross-modal reasoning in MLLMs.
		% Specifically, RAP identifies cognitive samples by evaluating each sample's intrinsic potential to activate genuine multi-modal reasoning by two complementary estimators:(1) a \textit{Causal Divergence Estimator (CDE)} based on the Potential Outcome Model pricinple, capturing samples that overly rely on language priors by comparing output differences between multi-modal and text-only inputs; (2) an \textit{Attention Confidence Estimator (ACE)}, which exploits token-level self-attention distributions to remove samples dominated by irrelevant but overweighted tokens in intermediate reasoning stages.
		Building on this insight, we propose a novel data selection paradigm termed \textit{\textbf{R}easoning \textbf{A}ctivation \textbf{P}otential (RAP)}, which identifies cognitive samples by estimating each sample's potential to stimulate genuine multi-modal reasoning by two complementary estimators: 1) \textit{Causal Discrepancy Estimator (CDE)} based on the potential outcome model principle, eliminates samples that overly rely on language priors by comparing outputs between multi-modal and text-only inputs; 2) \textit{Attention Confidence Estimator (ACE)}, which exploits token-level self-attention to discard samples dominated by irrelevant but over-emphasized tokens in intermediate reasoning stages.
		Moreover, we introduce a \textit{Difficulty-aware Replacement Module (DRM)} to substitute trivial instances with cognitively challenging ones, thereby ensuring complexity for robust multi-modal reasoning.
		Experiments on six datasets show that our RAP method consistently achieves superior performance \textit{using only 9.3\% of the training data, while reducing computational costs by over 43\%}.
		%Our code is available at \url{https://github.com/Mrshenshen/RAP}.
	\end{abstract}
	

	\begin{figure}[t]
		\centering
		%\vspace{-1.5em}
		\includegraphics[width=0.475\textwidth]{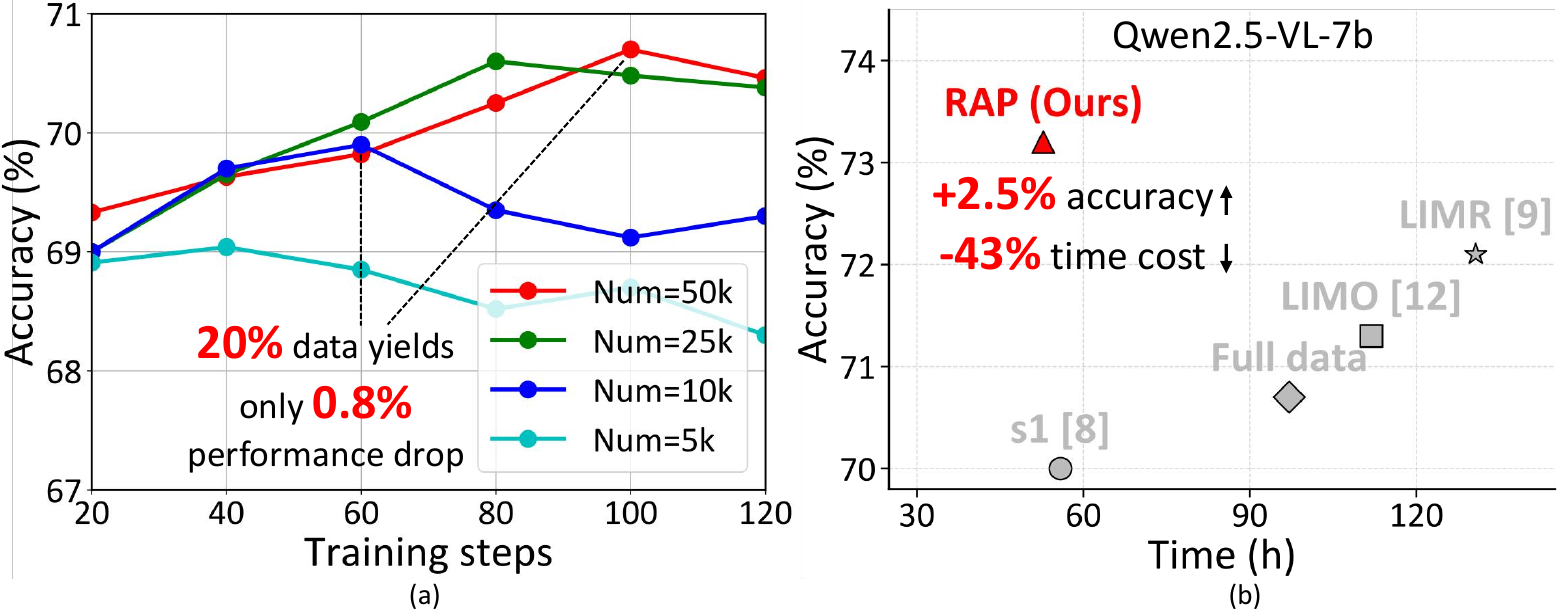}
		\vspace{-4mm}
		\caption{
			Comparison of (a) accuracy under varying training dataset sizes and (b) performance--efficiency trade-offs on various methods.
		}
		\label{fig:intro1}
		%\vspace{-4mm}
	\end{figure}
	
	\section{Introduction}
	\IEEEPARstart{I}mproving complex reasoning in multi-modal large language models (MLLMs) \cite{qwen2025qwen25technicalreport, openai2023gpt4} remains a fundamental challenge.
	%which is essential for models to address real-world problems that require sophisticated reasoning.
	While large-scale reinforcement learning (RL) \cite{guo2025deepseek, hu2024openrlhf, r1vDBLP:journals/corr/abs-2503-10615} has shown promise in enhancing reasoning capability, the prevailing assumption \cite{openai2024reasoning, kimiteam2025kimik15scalingreinforcement} suggests that scaling training data is a necessary condition for developing advanced reasoning ability, thus leading to data redundancy and substantial training costs.
	%However, the relationship between data volume and reasoning ability remains largely unexplored. 
	% Recent studies have observed that curated high-quality data can outperform training on full datasets for unimodal reasoning, it remains unclear whether this "less is more" phenomenon extends to multi-modal settings, where effective integration across modalities is essential.
	Recent studies \cite{s1, limr} indicate that LLMs trained on high-quality curated datasets can outperform those trained on full corpora.
	However, it remains unclear whether this principle generalizes to multi-modal contexts, where effective cross-modal integration is important.
	% This raises a critical question: Does scaling RL data always enhance cross-modal reasoning in MLLMs?
	This raises a critical question: \textit{can smaller high-value data achieve competitive or superior multi-modal reasoning compared to training on full corpora in MLLMs?}
	% To investigate this, as shown in Figure \ref{fig:intro1}, we perform an empirical analysis of how data scale affects cross-modal reasoning.
	% Using 20\% of training data only bring 1\% performance drop compared to the full data.
	% Such results demonstrate only marginal improvements or even slight performance degradation as the data volume increases.
	To investigate this, as shown in Fig. \ref{fig:intro1}(a), we empirically analyze the effect of data scale on multi-modal reasoning performance. 
	Notably, \textit{training with only 20\% of the data leads to merely a 0.8\% performance degradation compared to the full dataset}, suggesting that indiscriminate data scaling may have minimal or even negative effects.
	We hypothesize that such data augmentation diminishes the influence of high-value samples, termed \textit{cognitive samples}, which are essential for guiding effective cross-modal integration during reasoning.
	% We hypothesize the reason is that indiscriminate data expansion may dilute the benefit from high-value samples, referred to as ``\textit{cognitive samples}'', which guide the model to integrate information across modalities to solve complex reasoning tasks.
	%We refer to such samples as cognitive samples, since they compel the model to integrate multi-modal information in a meaningful way to arrive at correct solutions.
	%We refer to these samples \textit{`cognitive samples''}, which require the model to integrate information across modalities to solve a task.
	
	\begin{figure*}[t]
		\centering
		\includegraphics[width=0.99\linewidth]{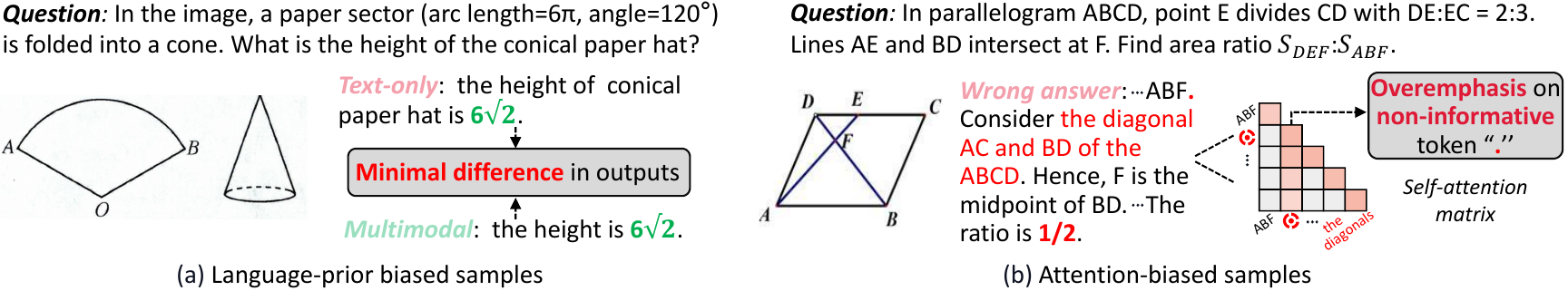}
		\caption{Illustrative examples for two ineffective training sample types: (a) language-prior biased samples and (b) attention-biased samples.}
		\label{fig:intro2}
		%\vspace{-8mm}
	\end{figure*}

	To validate this assumption, we analyze the characteristics of training samples and find that most fail to encourage joint attention to both modalities during reasoning.  
	Specifically, we identify two main types of ineffective samples: 1) \textit{Language-prior biased samples} (Fig. \ref{fig:intro2}(a)), where the model produces nearly indistinguishable outputs given text-only and multi-modal inputs due to over-reliance on language priors \cite{lp1, vcd}. 
	Such samples enable models to solve tasks with minimal utilization of visual semantics, thus impairing their ability to cross-modal integration.
	(2) \textit{Attention-biased samples} (Fig. \ref{fig:intro2}(b)), where the model over-attends to semantically irrelevant tokens (\egs, the punctuation ``.''), thereby obstructing the exploration of crucial cross-modal relationships.   
	%These samples hinder the model from exploring essential cross-modal associations during training.
	% While these issues highlight the need for selective data usage, existing data-effcient methods have primarily focused on selecting textual samples based on difficulty estimation \cite{} or reward-driven sampling strategies \cite{}. 
	% However, such approaches can not assess the extent to which individual samples contribute to the development of cross-modal reasoning by promoting multi-modal integration.
	% These findings highlight the necessity for data selection methods in multi-modal RL that extend beyond sample quality and explicitly consider cross-modal interactions.
	% However, existing methods have mainly focused on unimodal textual quality, relying on difficulty estimation \cite{} or reward-based sampling \cite{} to guide selection. Such methods fails to evaluate whether a given sample facilitates the model’s ability to integrate information across modalities.
	% These samples hinder the model from exploring crucial cross-modal relationships during training. 
	The above findings highlight the need to prioritize cross-modal interactions in data selection for multi-modal reasoning.
	% However, existing methods that rely on unimodal textual quality, such as human-annotated difficulty estimation \cite{s1, limr} or reward-based sampling \cite{limo}, entail significant manual annotation costs \cite{s1, limo} and extended filtering time \cite{limr}, fail to assess the ability of samples to enhance cross-modal integration.
	However, existing methods rely on unimodal textual quality, such as human-annotated difficulty estimation \cite{s1, limr} or reward-based sampling \cite{limo}.
	These approaches not only incur substantial manual annotation costs \cite{s1, limo} and considerable filtering time \cite{limr}, but also fail to estimate whether samples effectively facilitate cross-modal integration.

	Motivated by the above observations, we propose a novel data selection paradigm termed \textit{\textbf{R}easoning \textbf{A}ctivation \textbf{P}otential (RAP)} for enhancing multi-modal reasoning while reducing training costs.
	RAP aims to identify cognitive samples that effectively trigger multi-modal reasoning during RL training.
	Specifically, RAP estimates the reasoning potential of each sample through two complementary perspectives: \textit{output-level reasoning discrepancy} and \textit{process-level reasoning confidence}.
	For the former, we are inspired by the intuition that if model predictions remain invariant regardless of visual input presence, the model may merely exploit linguistic biases rather than engaging in  genuine multi-modal reasoning. 
	We formalize this notion through \textit{Causal Discrepancy Estimator (CDE)}, which employs the Potential Outcome Model (POM) to estimate the causal effect of input modality on model predictions by simulating counterfactual outcomes, \ies, what the model would output if one modality were removed. 
	% Based on this principle, the CDE quantifies the output discrepancy between multi-modal and text-only inputs to eliminate \textit{language-prior biased samples}.
	Consequently, CDE effectively eliminates \textit{language-prior biased samples} by measuring discrepancies between multi-modal and text-only predictions.
	
	However, relying solely on output-level measures neglects the reliability of internal reasoning dynamics.
	Therefore, we propose the \textit{Attention Confidence Estimator (ACE)} to model the quality of internal reasoning behavior based on token-level attention distributions, thus removing \textit{attention-biased samples} characterized by high attention to irrelevant tokens. 
	%These two estimators collectively allow RAP to identify cognitive samples, thereby achieving efficient cross-modal reasoning in multi-modal RL.
	Despite their efficacy, combining these two estimators alone may retain overly simplistic samples while discarding challenging yet valuable instances, thereby constraining the model's reasoning upper bound. 
	To address this limitation, we propose a \textit{Difficulty-aware Replacement Module (DRM)} to replace trivial samples with suitable challenging alternatives, which ensure sufficient data complexity for robust multi-modal reasoning.
	Finally, the results in Fig. \ref{fig:intro1}(b) demonstrate that our RAP method \textit{achieves state-of-the-art performance with only 5,159 samples, compared to the full dataset of 54,931 samples, while reducing training costs by over 43\%.}
	% These findings validate our insight that data quality outweighs blind data scaling for multi-modal reasoning in RL, revealing the ``\textit{truth in the few}'', a viable direction for future work in efficient cross-modal reasoning.
	These findings validate our insight that data quality is more important than blind data scaling for multi-modal reasoning in RL, revealing the ``\textit{truth in the few}'' phenomenon.
	
	% \textbf{Our main contribution:} 1) We propose a novel \textit{Reasoning Activation Potential (RAP)} data selection framework to identify cognitive samples for reducing training cost and enhancing reasoning in MLLM. 
	% 2) We design a \textit{Causal Discrepancy Estimator (CDE)} to model the causal effects of different modalities via POM, quantifying output discrepancy under multi-modal and text-only inputs to remove samples overly driven by language priors.
	% 3) We introduce a \textit{Attention Confidence Estimator (ACE)} to eliminate attention-biased samples, where the model excessively focuses on irrelevant tokens.
	% 4) We propose a \textit{Difficulty-aware Replacement Module (DRM)} to ensure sufficient data complexity for improving reasoning upper bound.
	
	% \textbf{Our main contribution:} 1) We propose a novel \textit{Reasoning Activation Potential (RAP)} data selection paradigm, which identifies cognitive samples to reduce training costs and improve reasoning in MLLMs. 
	% 2) We develop a \textit{Causal Discrepancy Estimator (CDE)} to filter language-prior-biased samples and an \textit{Attention Confidence Estimator (ACE)} to remove attention-biased samples.
	% 3) We introduce a \textit{Difficulty-aware Replacement Module} to ensure data complexity by replacing trivial samples with challenging alternatives, thereby maximizing the model’s reasoning upper bound.
	
	Overall, our contributions can be summarized as follows:
	\begin{itemize}
		% \item We propose a novel framework named \textit{Cross-modal Uncertainty Modeling with Diffusion-based Refinement (CUMDR)} to alleviate the overfitting issue by incorporating uncertainty modeling and learnt joint probabilities.
		\item We reveal a ``\textit{truth in the few}'' phenomenon that smaller high-quality datasets can outperform full corpora for multi-modal reasoning in LVLMs.
		\item We propose two novel estimators: a \textit{Causal Discrepancy Estimator (CDE)} to eliminate samples that overly rely on language priors, and an \textit{Attention Confidence Estimator (ACE)} to filter out attention-biased samples with irrelevant semantic focus.
		\item We introduce a \textit{Difficulty-aware Replacement Module (DRM)} to preserve sufficient data complexity, effectively improving the model’s reasoning performance ceiling.
		% \item We achieve new state-of-the-art performance on all evaluation metrics of three datasets, and extensive experiments conducted on three benchmark datasets verify the superiority of our model.
		%	\item Extensive experiments on all benchmark datasets of ZS-SBIR verify the effectiveness of our TVT superiority.
	\end{itemize}
	% 4) Experiments demonstrate state-of-the-art performance with only 5,159 samples compared to the 54,931 in the full dataset, significantly reducing training costs by over 46%.
	% 1) We propose and demonstrate a ``\textit{truth in the few}'' hypothesis that scaling RL training data is not necessarily for better reasoning performance in MLLMs. 
	% 2) We design a novel \textit{Reasoning Activation Potential (RAP)} data selection paradigm to identify cognitive samples for reducing training costs and enhancing reasoning in MLLMs. 

	% . guided by the Potential Outcome Model (POM) principle, 
	% and (2) intermediate reasoning confidence, measured via token-level attention dynamics, to filter out samples dominated by irrelevant or misleading cues.
	
	\begin{figure*}[t]
		\centering
		\includegraphics[width=0.99\linewidth]{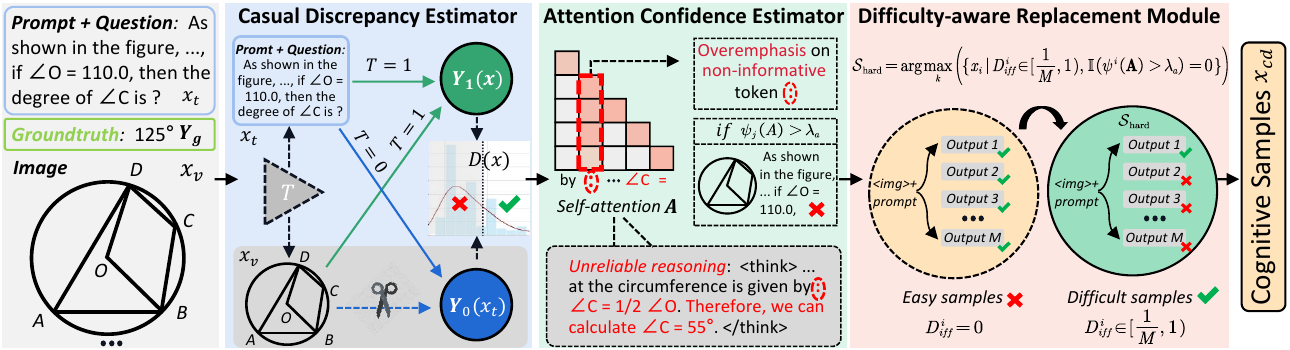}
		\caption{The overall pipeline of our RAP method. First, the Causal Discrepancy Estimator (CDE) filters out samples that overly rely on language priors via output-level discrepancy. Then, the Attention Confidence Estimator (ACE) excludes attention-biased samples by token-level attention distributions. Finally, the Difficulty-aware Replacement Module (DRM) selectively replaces trivial instances with cognitively challenging ones, yielding a refined subset of cognitive samples.}
		\label{fig:fra}
		%\vspace{-6mm}
	\end{figure*}
	
	\section{Related Work}
	\subsection{Reinforcement Learning for (M)LLM Reasoning}
	Reinforcement learning (RL), initially popularized by RL from Human Feedback (RLHF) \cite{hu2024openrlhf, rlhf1}, has become a key paradigm for improving the reasoning ability of both LLMs \cite{tmm1} and MLLMs \cite{tmm2, tmm3}. 
	Recent methods \cite{step-dpo, PR-DPO, dpo} extend RL beyond human preference alignment, explicitly achieving reasoning improvement via policy-gradient algorithms such as Proximal Policy Optimization (PPO) \cite{ppo} and reward-centric optimizations including Reinforce Leave-One-Out (RLOO) \cite{rloo} and Group Relative Policy Optimization (GRPO) \cite{guo2025deepseek}.
	Moreover, several methods \cite{lmmr1, r1vDBLP:journals/corr/abs-2503-10615, vrftliu2025visual, mpowang2024enhancing} explore the use of RL to enhance the visual reasoning in MLLMs.
	Recent works have extended RL beyond alignment for enhancing reasoning capability, by utilizing policy gradient algorithms PPO \cite{ppo} and optimizing reward signals, such as RLOO \cite{rloo} and GRPO \cite{guo2025deepseek}.
	However, these methods typically rely on large-scale data \cite{mulberry, MM-Eureka}, which fails to adequately consider the quality of training samples. 
	To this end, we introduce a novel RAP method to select valuable samples, ensuring that the training process stimulates effective multi-modal reasoning.

	\subsection{Data Selection for Reasoning}
	Traditional approaches \cite{guo2025deepseek, kimiteam2025kimik15scalingreinforcement, mulberry, TMM-lss} generally highlight the importance of data scaling, suggesting that larger data volumes can lead to better model performance.
	%Contrary to this, recent methods \cite{limo, limr, s1} demonstrate that curated datasets can outperform those trained on full corpora in complex textual reasoning tasks. 
	In contrast, recent methods \cite{limo, limr, s1} demonstrate that carefully curated datasets can outperform models trained on full corpora for challenging textual reasoning tasks. 
	For example, LIMO \cite{limo} and s1 \cite{s1} show that models trained with only 817 or 1000 curated samples can achieve better performance than those trained with substantially larger datasets, indicating that sample quality and representativeness may be more decisive than sheer quantity. 
	Similarly, LIMR \cite{limr} suggests that selective data curation during RL training can further improve outcomes with fewer samples by prioritizing instances that provide clearer optimization signals.  
	Inspired by these, we focus on whether and how a minimal but valuable dataset can similarly enhance multi-modal reasoning within multi-modal contexts.
	
	\subsection{Causal Mechanism}
	Recent advances have highlighted the role of causal mechanisms in understanding and improving complex reasoning \cite{casualnipsDBLP:conf/nips/RohekarGN23, casualrw2DBLP:conf/aaai/ZhangZZ24, r4_tmm2025, tim1}. 
	To capture causal relationships in MLLM reasoning, prior studies have leveraged established causal inference frameworks, such as Structural Causal Models (SCM) \cite{scholkopf2012causal} and Potential Outcome Models (POM) \cite{pom}. 
	Compared with purely correlational analyses, these frameworks facilitate intervention-based comparisons that help isolate factors genuinely responsible for a model's decisions. 
	In particular, POM provides a rigorous analytical structure to examine causal dependencies \cite{casualnipsDBLP:conf/nips/RohekarGN23, pom-llm}, enabling the quantification of how an explicit change in input conditions affects model outputs. 
	Therefore, we adopt the POM framework to systematically measure differences in MLLM predictions under multi-modal inputs versus text-only inputs, thereby assessing the causal contribution of visual information to the reasoning process.

	\section{Method}
	\noindent\textbf{Overview of RAP Method.}
	As shown in Fig. \ref{fig:fra}, our RAP method aims to identify high-value training samples $x_{cd}$, termed cognitive samples, that effectively stimulate multi-modal reasoning in MLLMs during RL training. 
	Given a training instance $x = (x_{t}, x_v)$, we estimate its reasoning activation potential from two perspectives. 
	First, we adopt the potential outcome model to quantify the output-level discrepancy $D$ between the model predictions under multi-modal inputs, $\textit{\textbf{Y}}_1(x)$, and text-only inputs, $\textit{\textbf{Y}}_0(x_{t})$. 
	Samples with low discrepancy values be considered as language-prior biased and are discarded accordingly.
	Second, we compute a confidence score $\psi(\textit{\textbf{A}})$ from the self-attention matrix $\textit{\textbf{A}} \in \mathbb{R}^{d \times d}$ of the model's final layer to assess the model’s focus on semantically meaningful tokens. 
	Samples with attention focused on irrelevant tokens, below a threshold $\lambda_a$, are excluded as attention-biased.
	Moreover, we replace overly easy samples with an equal number of hard examples potentially missed by the initial model due to limited reasoning capacity, thereby enhancing the model’s reasoning upper bound.
	The resulting cognitive samples form a refined training dataset that supports more efficient and robust multi-modal reasoning in MLLMs.
	Finally, we utilize these cognitive samples $x_{cd}$ to optimize the model by maximizing the objective of GRPO \cite{guo2025deepseek, hintgrpo}.
	% \begin{align}
		% \mathcal{J}_{GRPO}(\theta) &= \mathbb{E}[q \sim P(Q), \{o_i\}_i^G \sim \pi_{\theta_{\text{old}}}(O|q)] \\ \nonumber
		% &\quad \frac{1}{G} \sum_{i=1}^{G} \frac{1}{|o_i|} \sum_{t=1}^{|o_i|} \left\{
		%     \pi_{\theta}(o_{i,t}|q, o_i, \dots) \hat{A}_{i,t} - \beta D_{\text{KL}}[\pi_{\theta_0} \parallel \pi_{\text{ref}}] \right\}.
		% \end{align}

	%The resulting cognitive samples form a refined training dataset that supports more efficient and robust cross-modal reasoning in MLLMs.
	%The output is a curated dataset that enables efficient reinforcement learning with reduced computational cost and enhanced reasoning performance.
	
	\begin{algorithm}[t]
		\caption{The Process of Reasoning Activation Potential (RAP) Method.}
		\label{alg:rap}
		\begin{algorithmic}[1]
			\Require Original dataset $D=\{(x_t^i,x_v^i,y^i)\}_{i=1}^N$, pretrained model $\theta$, output discrepancy threshold $\lambda_c$, attention confidence threshold $\lambda_a$, training steps $T$
			\Ensure Curated cognitive samples $x_{cd}$
			
			\Statex \hspace{-0.6em}\textbf{Stage 1: Causal Discrepancy Estimator (CDE)}
			\ForAll{$(x_t,x_v,y)\in D$}
			\State Generate $M$ outputs with multimodal input $(x_t,x_v)$ and text-only input $(x_t)$
			\State Compute output consistency:
			\State \hspace{1.2em}$D(x)=\frac{1}{M}\sum_{j=1}^M\Big[\mathbb{I}\!\big(Y_1^{(j)}=y\big)-\mathbb{I}\!\big(Y_0^{(j)}=y\big)\Big]$
			\EndFor
			\State Form discrepancy-filtered set:
			\State \hspace{1.2em}$X_{\mathrm{cde}}=\{\,x\mid D(x)\ge \mu_D+\lambda_c\sigma_D\,\}$
			
			\Statex \hspace{-0.6em}\textbf{Stage 2: Attention Confidence Estimator (ACE)}
			\State Initialize $X_{\mathrm{ace}}\gets\emptyset$
			\ForAll{$x\in X_{\mathrm{cde}}$}
			\State Run a forward pass and extract final-layer self-attention $A\in\mathbb{R}^{L\times L}$
			\State Compute attention confidence:
			\State \hspace{1.2em}$\psi_j(A)=\prod_{i=j}^L(\sigma\cdot A_{i,j}),\quad \psi(A)=\max_j \psi_j(A)$
			\If{$\psi(A)\le \lambda_a$}
			\State $X_{\mathrm{ace}}\gets X_{\mathrm{ace}}\cup\{x\}$
			\EndIf
			\EndFor
			
			\Statex \hspace{-0.6em}\textbf{Stage 3: Difficulty-aware Replacement Module (DRM)}
			\ForAll{$x\in X_{\mathrm{ace}}$}
			\State Compute difficulty:
			\State \hspace{1.2em}$D_{\mathrm{diff}}(x)=1-\frac{1}{M}\sum_{j=1}^M \mathbb{I}\!\big(\text{output}^{(j)}\text{ is correct}\big)$
			\EndFor
			\State Identify trivial samples:
			\State \hspace{1.2em}$X_{\mathrm{easy}}=\{\,x\mid D_{\mathrm{diff}}(x)=0\,\}$
			\State Select top $|X_{\mathrm{easy}}|$ challenging samples from $D\setminus X_{\mathrm{ace}}$ by descending difficulty to form $X_{\mathrm{hard}}$
			\State Construct final cognitive dataset:
			\State \hspace{1.2em}$x_{cd}=(X_{\mathrm{ace}}\setminus X_{\mathrm{easy}})\cup X_{\mathrm{hard}}$
			
			\Statex \hspace{-0.6em}\textbf{Stage 4: Cognitive Dataset Training}
			\State Train $\theta$ on $x_{cd}$ via reinforcement learning algorithm (GRPO) for $T$ steps
			
		\end{algorithmic}
	\end{algorithm}

	\subsection{Causal Discrepancy Estimator}
	% xxxx(开头介绍)
	% \noindent\textbf{Background of Potential Outcome Model.} xxxx
	% \noindent\textbf{Output-level Divergence Calculation.} xxxx
	To identify samples where the model genuinely engages in multi-modal reasoning, rather than overly relying on language priors, we interpret modality influence in reasoning as output discrepancy between multi-modal and text-only inputs, formulated under the Potential Outcome Model (POM).
	
	% \noindent\textbf{Background of Potential Outcome Model.}
	% The foundation of causal inference is rooted in the Neyman-Rubin Potential Outcome Model (POM) \cite{rubin1974estimating}, which seeks to know the effect of a treatment $T$ on the outcome $Y$ for individuals described by covariates .
	% In this work, we consider the random variable $T$ to be binary ($\gamma$ = {0, 1}).
	% Under this framework, each unit $u$ is associated with two potential outcomes: $Y_1(u)$, the outcome if the unit receives the treatment ($T=1$), and $Y_0(u)$, the outcome under control ($T=0$). 
	% The Individual Treatment Effect (ITE) is defined as the difference $Y_1(u) - Y_0(u)$. However, due to the fundamental problem of causal inference \cite{holland1986statistics}, only one of these outcomes can be observed for any given unit, rendering the ITE inherently unidentifiable in practice.
	
	% To mitigate this, \cite{} propose the \emph{Conditional Average Treatment Effect} (CATE), which captures the expected effect of the treatment conditioned on units:
	% \[
	% \tau(x) = \mathbb{E}[Y_1 - Y_0 \mid X = x],
	% \]
	% where $x$ denotes the observed features of the unit. CATE serves as a more tractable target for estimation.
	
	\noindent\textbf{Background of Potential Outcome Model.} 
	The foundation of causal inference is rooted in the Neyman-Rubin POM \cite{rubin1974estimating, pom}, which aims to estimate the effect of a treatment $T$ on an outcome $\textit{\textbf{Y}}$ for individuals described by covariates $\textit{\textbf{X}}$. 
	In this work, we consider the treatment variable $T$ to be binary, \ies, $T \in \{0, 1\}$.  
	Under this framework, each unit $u$ is associated with two potential outcomes: $\textit{\textbf{Y}}_1(u)$, the outcome if the unit receives the treatment ($T=1$), and $\textit{\textbf{Y}}_0(u)$, the outcome under control ($T=0$). 
	The Individual Treatment Effect (ITE) \cite{ite1, ite2} is defined as the difference $\textit{\textbf{Y}}_1(u) - \textit{\textbf{Y}}_0(u)$. 
	However, due to the fundamental problem of causal inference \cite{itedrawpearl2009causal, itedraw2hammerton2021causal}, only one of these outcomes can ever be observed for a given individual, rendering the ITE fundamentally unidentifiable.
	To address this, the prior work \cite{CATEabrevaya2015estimating} proposes the Conditional Average Treatment Effect (CATE), which represents the expected treatment effect conditioned on covariates:
	\begin{equation}
		\mathbb{E}[\textit{\textbf{Y}}_1 - \textit{\textbf{Y}}_0 \mid \textit{\textbf{X}} = x] 
		= \mathbb{E}[\textit{\textbf{Y}} \mid T=1, \textit{\textbf{X}}=x] - \mathbb{E}[\textit{\textbf{Y}} \mid T=0, \textit{\textbf{X}}=x],
	\end{equation}
	where $x$ denotes the observed covariates of the unit. 
	%The CATE provides a more tractable target compared to the ITE.  
	%A detailed theoretical justification can be found in Appendix.

	% Causal identification from observational data requires three standard assumptions:
	% \begin{itemize}
		%     \item \textbf{Consistency}: The observed outcome $Y$ matches the potential outcome under the actual treatment received, i.e., $Y = T \cdot Y_1 + (1 - T) \cdot Y_0$.
		%     \item \textbf{Unconfoundedness (Ignorability)}: Treatment assignment is independent of the potential outcomes given the covariates, i.e., $(Y_0, Y_1) \perp T \mid X$.
		%     \item \textbf{Overlap (Positivity)}: Every unit has a non-zero probability of receiving either treatment, i.e., $0 < P(T=1 \mid X=x) < 1$ for all $x$.
		% \end{itemize}
	% These assumptions enable the estimation of causal effects using statistical or machine learning models trained on observational datasets.

	\noindent\textbf{Output-level Discrepancy Calculation.}
	Inspired by the intuition that a model generating nearly identical outputs in the presence or absence of visual input may fail to use multi-modal information for reasoning, we employ the POM to formalize the influence of the visual modality on model predictions, by defining outcomes under distinct treatment conditions.
	% Specifically, given an input $x = (x_{\text{text}}, x_{\text{image}})$, we define the presence of the image as the binary treatment variable $T \in \{0, 1\}$.
	% Then, we can define two potential outcomes, where $Y_1(x_{\text{text}})$ denotes the model output when exposed to both text and image modalities.
	% $Y_0(x_{\text{text}})$ represent the counterfactual output when only text modality is provided.
	Specifically, we treat the presence of the image as a binary treatment variable \( T \in \{0, 1\} \), where \( T = 1 \) indicates the inclusion of the image and \( T = 0 \) denotes its absence.  
	Given an input \( x = (x_{t}, x_{v}) \), we define two potential outcomes: \( \textit{\textbf{Y}}_1(x) \), the model's output given both text and image, and \( \textit{\textbf{Y}}_0(x_{t}) \), the counterfactual output when only the text is provided.  
	The text-only output \( \textit{\textbf{Y}}_0(x_{t}) \) can be calculated as follows:
	\begin{equation}
		y_i \in \textit{\textbf{Y}}_0(x_t) \sim \text{softmax} \left[ \log_{\theta} \left( y_i \mid x_t\right) \right], 
	\end{equation}
	
	The multi-modal output \( \textit{\textbf{Y}}_1(x) \) can be calculated as follows:
	\begin{equation}
		y_i \in \textit{\textbf{Y}}_1(x) \sim \text{softmax} \left[ \log_{\theta} \left( y_i \mid x_v, x_t\right) \right], 
	\end{equation}
	
	To quantify the discrepancy between model outputs $\textit{\textbf{Y}}_0(x_t)$ and $\textit{\textbf{Y}}_1(x)$ under multi-modal and text-only inputs, we compute the consistency of these model outputs with the ground truth $\textit{\textbf{Y}}_g$.  
	If the model's output matches the ground truth in a given condition, we assign a value of 1; otherwise, a value of 0. 
	The discrepancy $D(x)$ for each sample $x$ is then quantified as the normalized difference in the number of correct predictions between these conditions, which can be formulated as:
	\begin{equation}
		\begin{aligned}
			D(x) &= \mathbb{E}\!\left[ \mathbb{I}(\textit{\textbf{Y}}_1 = \textit{\textbf{Y}}_g) - \mathbb{I}(\textit{\textbf{Y}}_0 = \textit{\textbf{Y}}_g) \mid x \right] \\
			&= \frac{1}{M} \sum_{i=1}^{M} \left[ \mathbb{I}\!\big(Y_1(x^{(i)}) = Y_g^{(i)}\big) - \mathbb{I}\!\big(Y_0(x_t^{(i)}) = Y_g^{(i)}\big) \right],
		\end{aligned}
	\end{equation}
	where \( M \) is the number of rollout outputs generated for each sample set in GRPO, and \( \mathbb{I}(\cdot) \) is the indicator function that equals 1 if the condition is true, and 0 otherwise.
	% To further refine the selection of samples, we set a threshold for the divergence score based on the mean and standard deviation of divergences across all samples. Specifically, samples with divergence smaller than \( \text{mean} - \lambda \times \text{std} \), where \( \lambda \) is a hyperparameter, are excluded from the training set.
	% This procedure effectively filters out samples where the addition of the image modality does not provide substantial improvement over the text modality, thus selecting the most informative samples for cross-modal reasoning.
	Based on the mean and standard deviation of discrepancies across all samples, we set a threshold for the discrepancy score. 
	Specifically, samples with discrepancy less than \( \mu_c + \lambda_c \cdot \sigma_c \), where \( \mu_c \) is the mean discrepancy, \( \sigma_c \) is the standard deviation, and \( \lambda_c \) is a tunable hyperparameter, are excluded from the training set.  

	\subsection{Attention Confidence Estimator}
	\label{sub:3.2 ace}
	While the CDE effectively identifies samples requiring multi-modal reasoning from an output-level perspective, it does not assess the quality of internal reasoning processes. 
	Recent studies \cite{opera24, anctokDBLP:conf/emnlp/WangLDCZMZS23} reveal an insightful phenomenon: tokens that receive excessive attention weights can dominate the prediction without using a meaningful semantic context. 
	Motivated by this, we explicitly quantify the internal reasoning quality via self-attention distributions, thus filtering out \textit{attention-biased samples}.
	
	\noindent\textbf{Attention Confidence Formulation.}  
	Given an input $x = (x_t, x_v)$ to a transformer-based MLLM, we denote the self-attention matrix $\textit{\textbf{A}} \in \mathbb{R}^{d \times d}$ from its final transformer layer as:
	% \begin{equation}
		% A \in \mathbb{R}^{L \times L}, \quad A_{i,j} = \text{softmax}\left(\frac{Q_i K_j^\top}{\sqrt{d}}\right),
		% \end{equation}
	\begin{equation}
		\quad A_{i,j} = \text{softmax}\left(\frac{\textit{\textbf{Q}}_i \textit{\textbf{K}}_j^\top}{\sqrt{d}}\right),
	\end{equation}
	where $\textit{\textbf{Q}}_i, \textit{\textbf{K}}_j \in \mathbb{R}^{d}$ represent the query and key vectors for token positions $i$ and $j$.
	
	To systematically characterize attention-bias patterns, ACE analyzes the entire self-attention matrix \( \textit{\textbf{A}} \). 
	An attention-biased pattern at token position \( j \) is identified if the corresponding attention column exhibits a pronounced concentration of attention weights, identifying excessive reliance on a single token. 
	Formally, the degree of attention bias at position \( j \), $\psi_j(\textit{\textbf{A}})$ is quantified by computing a multiplicative attention score across subsequent token interactions:
	\begin{equation}
		\psi_j(\textit{\textbf{A}}) = \prod_{i=j}^{L}(\sigma \cdot A_{i,j}),
	\end{equation}
	where \(\sigma\) is a scaling factor ensuring numerical stability and emphasizing prominent attention patterns.  
	$L$ denotes the total length of the input sequence.
	The \( \psi_j(A) \) metric effectively quantifies attention confidence, with elevated values indicating unreliable multi-modal reasoning. 
	We define a position \( j \) as attention-biased if \(\psi_j(A)\) exceeds a threshold \(\lambda_a\). 
	Instances containing more than one attention-biased position are filtered out as attention-biased samples.
	% The total aggregation pattern count \( Z(A) \) across the self-attention matrix \( A \) is then computed as, which can effectively reflect the attention confidence:
	% \begin{equation}
		% Z(A) = \sum_{j=1}^{L}\mathbb{I}\left[\psi_j(A)>\gamma\right],
		% \end{equation}
	%where \(\mathbb{I}[\cdot]\) is an indicator function yielding 1 when the condition holds, and 0 otherwise.

	% \noindent\textbf{Threshold-based Attention Filtering.}  
	% To systematically filter attention-biased samples, ACE introduces a threshold-based selection criterion. Samples whose attention confidence exceeds a predefined threshold $\lambda_c$ (calculated empirically across training data) are considered attention-biased and thus excluded from training:
	% \begin{equation}
		% C(x) = 
		% \begin{cases}
			%     1, & \text{if } \phi(W_t^k) \leq \lambda_c, \\
			%     0, & \text{otherwise},
			% \end{cases}
		% \end{equation}
	% where $C(x)$ indicates whether a sample passes the attention confidence evaluation.
	
	\begin{table*}[t]
		\centering
		\small
		\caption{Comparison with state-of-the-art methods. Experiments are conducted using the Qwen2.5-VL-3b \cite{qwen2025qwen25technicalreport} and Qwen2.5-VL-7b \cite{qwen2025qwen25technicalreport}, employing GRPO as the RL method. ``Time'' denotes the total computation cost, including data selection and training. Bold font denotes the best result.}
		\setlength{\tabcolsep}{1.6pt}
		\setlength{\arrayrulewidth}{0.6pt}
		\begin{tabular}{l|c|cccccccc}
			\hline
			Method & Sample & \makecell{Time (h)} $\downarrow$ & \makecell{MathVista} $\uparrow$ & \makecell{MMStar} $\uparrow$ & \makecell{MathVerse} $\uparrow$ & \makecell{WeMath} $\uparrow$ & \makecell{MMVet} $\uparrow$ & \makecell{LogicVista} $\uparrow$ & Avg. $\uparrow$ \\
			%\hline
			%\multicolumn{10}{c}{Qwen2.5-VL-7B} \\
			\hline
			Qwen2.5-VL-7b & - & -  & 68.70 & 56.07 & 39.31 & 35.90 & 59.13 & 44.52 & 50.61 \\
			Qwen2.5-VL-7b-Full & 54,931 & 93.2 & 70.70 & 61.53 & 48.43 & 38.67 & 60.51  & 46.09 & 54.32 \\
			%Qwen2.5-VL-7b-Random & 5,159 & 42.7 & 69.10 & 59.42 & 42.17 & 34.62 & 59.32 & 44.41 & 51.51 \\
			Qwen2.5-VL-7b-s1 \cite{s1} (2025) & 1,000 & 55.9 & 68.50 & 61.80  & 45.79& 35.05 & 61.05 & 45.86 & 53.01 \\
			Qwen2.5-VL-7b-LIMO \cite{limo} (2025) & 4,093 & 111.9 & 69.90 & 61.33  & 45.74 & 34.67 & 59.08 & 44.74 & 52.58 \\
			Qwen2.5-VL-7b-LIMR \cite{limr} (2025) & 8,136 & 122.0 & 71.10  & 62.12 & 48.02 & 41.21 & 62.86 & 45.81 & 55.19 \\
			\rowcolor[gray]{0.94}
			\textbf{Qwen2.5-VL-7b-RAP (Ours)} & 5,159 & \textbf{52.8} & \textbf{73.20} & \textbf{62.53}  & \textbf{48.65} & \textbf{42.00} & \textbf{63.31} & \textbf{46.53} &\textbf{56.04} \\
			%Debiased Hint-GRPO & 46.68 & 54.19 & 45.69 & 32.18 & 14.99 & 14.61 & 45.86 & 38.55 \\
			%\hline
			%\multicolumn{10}{c}{Qwen2.5-VL-3B} \\
			\hline
			Qwen2.5-VL-3b & - & - & 61.30 & 54.46 & 9.01 & 21.62 & 51.81 & 39.59 &39.63 \\
			Qwen2.5-VL-3b-Full & 54,931 & 46.5 & 64.50 & 55.25 & 38.35 & 28.29 & 53.41 & 40.03 &46.64 \\
			%Qwen2.5-VL-3b-Random & 4,374 & 23.1 & 62.20 & 54.27 & 35.17 & 24.75 & 52.92 & 39.23 & 44.76 \\
			Qwen2.5-VL-3b-s1 \cite{s1} (2025) & 1,000 & 38.4 & 62.60 & 54.53  & 37.41 & 27.52 & 53.02 & 39.59 & 45.78 \\
			Qwen2.5-VL-3b-LIMO \cite{limo} (2025) & 2,679 & 120.5 & 61.80 & 54.73  & 35.33 & 24.48 & 53.12 & 39.14 & 44.77 \\
			Qwen2.5-VL-3b-LIMR \cite{limr} (2025) & 21,303 & 60.9 & 63.10 & 54.66 & 35.43  & 26.76 & 53.25 & 40.95 & 45.69  \\
			\rowcolor[gray]{0.94}
			\textbf{Qwen2.5-VL-3b-RAP (Ours)} & 4,374 & \textbf{32.0} & \textbf{64.90} & \textbf{55.67} & \textbf{39.34} & \textbf{29.33} & \textbf{54.63} & \textbf{41.61} & \textbf{47.58}\\
			\hline
		\end{tabular}
		\label{tab:main}
		%\vspace{-6mm}
	\end{table*}
	
	\subsection{Difficulty-aware Replacement Module}
	
	Although CDE and ACE select valuable samples, they inevitably limit the reasoning upper bound due to the exclusion of challenging yet informative samples.
	For example, in the CDE selection process, if the output discrepancy threshold is set above 0.2, a scenario may arise where a text-only model consistently produces incorrect outputs across all five trials, whereas multi-modal outputs succeed once in five trials. 
	Such challenging yet valuable samples, which could significantly contribute to improving reasoning, are thus discarded. 
	This exclusionary process reduces the complexity of the training data, thereby constraining the model to achieve more complex reasoning ability.

	To address this, we introduce a Difficulty-aware Replacement Module (DRM) to refine the selected sample.
	First, we define the difficulty score \( D_{iff}^i \) to quantify the challenge of correctly answering a sample, which can be defined as:
	\begin{equation}
		D_{iff}^i = 1 - \frac{\sum_{j=1}^{M} c_{i,j}}{M},
	\end{equation}
	where \( c_{i,j} \) denotes the correctness of the \( j \)-th rollout generation for the \( i \)-th sample, and \( M \) is the total number of outputs in the group. 
	A higher \( D_{iff}^i \) indicates greater difficulty. 
	% Moreover, we propose the deviation \( \Delta_i \) to measure the degree of answers deviates from the threshold \(\mu_c + \lambda_c \cdot \sigma_c\) set in CDE:
	% \begin{equation}
		%     \Delta_i = \left| \frac{\sum_{j=1}^{M} c_{i,j}}{M} - \lambda \right|,
		% \end{equation}
	% This helps identify samples near the threshold that may have been excluded due to marginal performance. 
	In particular, the DRM involves two steps: 
	First, we exclude easy samples, denoted as \( D_{iff}^i = 0 \), which are characterized by consistent correct answers across all trials.  
	Second, we reintroduce hard samples that have been previously discarded due to difficulty but are still valuable for training. 
	% The top-\( K \) hardest samples are selected based on both their difficulty score \( D_i \) and their deviation \( \Delta_i \) from the threshold \( \lambda \). 
	Specifically, based on the difficulty score \( D_{iff}^i \), the set of reintroduced samples \( \mathcal{S}_{\text{hard}} \) is given by:
	% \begin{equation}
		%     \mathcal{S}_{\text{hard}} = \left\{ x_i \mid D_i \in [\frac{1}{M}, 1) \text{ and } \psi_j^i(\textit{\textbf{A}}) = 0 \right\}
		% \end{equation}
	\begin{equation}
		\mathcal{S}_{\text{hard}} = \text{argmax}_k \left( \{ x_i \mid D_{iff}^i \in [\frac{1}{M}, 1), \, \mathbb{I}(\psi^i(\textit{\textbf{A}}) > \lambda_a) = 0 \} \right),
	\end{equation}
	% \[
	% \mathcal{S}_{\text{hard}} = \{ i \mid D_i > \gamma \text{ and } \Delta_i \text{ is among the top } \text{num} \}
	% \]
	where \( k \) is the number of easy samples. 
	This formulation identifies the top-\( k \) most challenging samples according to the difficulty metric \( D_{iff}^i \), while excluding those that do not meet the specified criteria.
	Note that the hardest samples, \ies, $D_i=1$, would be neglected, as they are demonstrated to be meaningless for training \cite{hintgrpo}.
	This DRM can enhance the upper bound of the model's ability to handle complex tasks, without introducing data redundancy and training costs.
	
	Finally, by filtering through CDE and ACE and refining with DRM, we ensure that the model is trained with cognitive samples $x_{cd}$, thereby enhancing the multi-modal reasoning ability, while simultaneously reducing training costs and data redundancy.

	\section{Experiments}
	\label{headings}
	
	\noindent\textbf{Training dataset.}
	Main results in the Table \ref{tab:main} are based on models trained with the \textit{MM-Eureka} dataset \cite{MM-Eureka}, a high-quality multi-modal dataset for mathematical reasoning. 
	To further validate the generalization of our RAP method, we evaluate models trained on the subset of \textit{Mulberry-260k} dataset \cite{mulberry}, a multi-modal learning-to-reason-and-reflect dataset.

	\noindent \textbf{Evaluation.}
	Similar to \cite{mulberry, MM-Eureka}, we evaluate models on both mathematical and general multi-modal reasoning tasks using the $pass@1$ metric, where $pass@1$ measures the percentage of problems correctly solved on the first attempt, under a zero-shot setting.
	For mathematical reasoning, we assess the model's ability on four benchmark datasets: MathVista \cite{mathvista}, MMStar \cite{mmstar}, MathVerse \cite{mathverse}, and WeMath \cite{wemath}. 
	For universal multi-modal reasoning, we evaluate on MMVet \cite{mmvet} and LogicVista \cite{logicvista}.

	\noindent \textbf{Implementation details.}
	Following prior methods \cite{MM-Eureka, mulberry}, we conduct our primary experiments on mathematical reasoning tasks, employing Qwen2.5-VL-3B and Qwen2.5-VL-7B \cite{qwen2025qwen25technicalreport} as baseline models.
	First, we apply RAP to select cognitive samples using the initial model without any training.
	These samples are then used to train models within the EasyR1 \cite{easyr1}, employing the AdamW optimizer with a learning rate of 1e-6.
	Full-data training requires 1 epoch (107 steps), all others use 40 training steps with a batch size of 512 across 8 GPUs. 
	The system prompt is available at our \textit{supplementary material}.
	% We also use DeepSpeed~\cite{aminabadi2022deepspeed, rajbhandari2020zero} to facilitate the model training through ZeRO-3 optimization.
	For accelerated generation in GRPO \cite{guo2025deepseek}, we utilize the vLLM package \cite{kwon2023efficient}. 
	Finally, we set the hyperparameters \(\sigma\) to 2.0, $\lambda_c$ and $\lambda_a$ to 0.5 and 0.1 for the CDE and ACE, respectively.

		\subsection{Overall Comparison Results}
		
		\noindent \textbf{Comparing methods.}
		We compare our approach with existing data selection methods, including: 1) \textit{s1} \cite{s1}, which utilizes the large-scale MLLMs to identify high-quality data; 2) \textit{LIMO} \cite{limo} that designs a difficulty-aware selection method to identify crucial samples; and 3) \textit{LIMR} \cite{limr}, which employs learning impact measurement to select a subset of training samples.
		Moreover, we also evaluate models trained on the full dataset (Full) as the baseline.

		\noindent \textbf{Comparisons with state-of-the-art methods.}
		We compare our RAP model with the latest data-efficient methods on six diverse datasets, including LIMO \cite{limo}, s1 \cite{s1} and LIMR \cite{limr}.
		The results shown in Table \ref{tab:main} reveal several key findings: 
		1) RAP consistently outperforms models trained with full corpora on all datasets. 
		% These improvements, achieved with only 9.5\% of the training data, support for our proposed "Truth in the Few" phenomenon. 
		% Specifically, training with a carefully curated subset of cognitive driver samples fosters more effective cross-modal reasoning compared to merely increasing the amount of RL data, while simultaneously reducing training time by 60\%.
		Remarkably, these improvements are achieved using only 9.5\% or 7.9\% of training data while reducing training time by 43\% or 31\%, supporting our hypothesis ``\textit{truth in the few}'' that selected cognitive samples can achieve more effective multi-modal reasoning. 
		2) Moreover, the RAP method demonstrates a significant improvement of 7.33\% and 6.95\% over the LIMO \cite{limo} and s1 \cite{s1} methods on WeMath \cite{wemath}, respectively, which rely on data quality and manual selection. 
		% Such results indicate that focusing on the potential of each sample to improve multi-modal reasoning, rather than focusing on the data itself, yields better performance in multi-modal scenarios.
		Such results indicate that focusing on the potential of each sample to improve multi-modal reasoning.
		
		\begin{table}[t]
			\centering
			\caption{Comparison with state-of-the-art methods using Qwen2.5-VL-7b \cite{qwen2025qwen25technicalreport} with the RL method RLOO \cite{rloo}, evaluating our RAP method on different training datasets and RL algorithms.}
			\label{tab:model-left}
			\small
			\setlength{\tabcolsep}{1.6pt}
			\begin{tabular}{l|cccc}
				\hline
				Method & MathVista & MMVet & We-Math & Avg.\\
				\hline
				Qwen2.5-VL-7b & 68.70  & 59.13  & 35.90 & 54.58   \\
				Qwen2.5-VL-7b-Full & 69.10 & 60.32 & 36.95 & 55.46  \\
				Qwen2.5-VL-7b-s1 \cite{s1}   & 68.50  & 59.96  & 35.06 & 54.51 \\
				Qwen2.5-VL-7b-LIMO \cite{limo} & 68.80  & 60.11  & 35.23 & 54.71 \\
				Qwen2.5-VL-7b-LIMR \cite{limr}   & 68.90  & 60.71  & 36.74 & 55.45 \\
				\textbf{Qwen2.5-VL-7b-RAP (Ours)} & \textbf{69.20} & \textbf{61.33} & \textbf{37.05} & \textbf{55.86}  \\
				\hline
			\end{tabular}
			%\vspace{-6mm}
		\end{table}
		
		\begin{table}[t]
			\centering
			\caption{Comparison with state-of-the-art methods using InternVL3-2b \cite{internvl} with the RL method GRPO \cite{guo2025deepseek}, evaluating our RAP method across different model architectures.}
			\label{tab:rloo-right}
			\small
			\setlength{\tabcolsep}{1.6pt}
			\begin{tabular}{l|cccc}
				\hline
				Method & MathVista & MMVet & We-Math & Avg.\\
				\hline
				InternVL3-2b & 56.10 & 58.22 & 12.06 & 42.13  \\
				InternVL3-2b-Full & 57.20 & 59.86 & 12.84 & 43.30  \\
				InternVL3-2b-s1 \cite{s1}   & 56.80 & 60.47 & 11.63 & 42.97 \\
				InternVL3-2b-LIMO \cite{limo} & 56.70 & 59.82 & 11.92 & 42.81  \\
				InternVL3-2b-LIMR \cite{limr}   & 57.10 & 61.33 & 12.44 & 43.62  \\
				\textbf{InternVL3-2b-RAP (Ours)} & \textbf{57.40} & \textbf{62.02} & \textbf{13.05} & \textbf{44.16}  \\
				\hline
			\end{tabular}
			%\vspace{-6mm}
		\end{table}

		\noindent \textbf{Effectiveness of RAP on different base models.} 
		As shown in Table \ref{tab:model-left}, our method consistently surpasses other recent data selection methods when applied to the base model InternVL3-2b \cite{internvl}. 
		% This outcome highlights the generalizability of the RAP method, as the proposed CDE and ACE can select training samples that activate the model's cross-modal reasoning, instead of exploiting architecture-specific inductive biases, thus making the RAP applicable to diverse model architectures
		This outcome highlights the broad applicability and generalizability of the RAP framework, as the introduced CDE and ACE components effectively select training samples that activate the model's multi-modal reasoning capability. 
		Crucially, these components do not rely on exploiting model-specific inductive biases, thereby ensuring RAP's adaptability on a wide range of model architectures.
		More results are available in our \textit{supplementary material}.

		\noindent \textbf{Effectiveness of RAP for various RL methods and training datasets.} 
		To further validate the generalizability of RAP, Table \ref{tab:rloo-right} presents results using the Qwen2.5-VL-7B model, trained under two different configurations: 1) the RLOO RL paradigm \cite{rloo}, and 2) the reduced Mulberry-10K dataset, a subset of Mulberry-260K \cite{mulberry}.  
		Despite these variations, RAP maintains consistent superiority compared to other methods, suggesting its generalization to different RL strategies and training datasets.
		We attribute this robustness to cognitive samples that facilitate genuine multi-modal reasoning rather than simply fitting data distributions.

		\begin{table}[t]
			\centering
			\small
			\setlength{\tabcolsep}{1.6pt}
			%\vspace{-3mm}
			\caption{Ablation study of RAP using Qwen2.5-VL-7B, trained using GRPO method for 40 steps.}
			\label{tab:ab-left}
			%\resizebox{\columnwidth}{!}{%
				\begin{tabular}{c|ccc|cccc}
					\hline
					No. & CDE & ACE & DRM & MathVista & MMStar & MathVerse & MMVet \\
					\hline
					0 & -& -&- & 69.10 & 59.32 & 46.07 & 58.91 \\
					1 & \ding{52} & & & 70.80 & 60.64 & 47.13 & 60.92 \\
					2 & & \ding{52} & & 70.20 & 60.21 & 46.86 & 60.68 \\
					3 & \ding{52} & & \ding{52} & 72.00 & 61.28 & 47.82 & 61.74 \\
					4 & & \ding{52} & \ding{52} & 71.50 & 60.93 & 47.90 & 61.33 \\
					5 & \ding{52} & \ding{52} & & 72.60 & 61.76 & \textbf{48.74} & 62.78 \\
					6 & \ding{52} & \ding{52} & \ding{52} & \textbf{73.20} & \textbf{62.53} & 48.65 & \textbf{63.31} \\
					\hline
				\end{tabular}%
			%\vspace{-5mm}
		\end{table}

		\subsection{Further Analysis}
		
		\noindent\textbf{Ablation study.}
		% To further investigate the significance of each module within our AUL method, we conducted ablation studies on various components, namely Uncertainty-aware Matching Filtration (UMF), Uncertainty-based Alignment Refinement (UAR), and Cross-modal Masked Modeling (CMM).
		As presented in Table \ref{tab:ab-left}, we list the following conclusions:
		1) The comparison between No.0 and No.3 indicates that integrating CDE and ACE can improve multi-modal reasoning in MLLMs. 
		These results underscore the efficacy of RAP in eliminating language-prior biased and attention-biased samples, thereby enabling models to focus on essential ``\textit{cognitive samples}'' and improving reasoning performance.
		2) Comparing No.1 with No.3 shows that only using the CDE can improve performance, but worse than the full RAP.
		We hypothesize the reason is that while CDE identifies critical samples by detecting output discrepancies, it overlooks the intermediate reasoning process. 
		%In contrast, ACE removes attention-biased data by estimating internal reasoning quality, thus enhancing the model’s reasoning ability.
		3) Moreover, the comparison between No.2 and No.3 demonstrates that the RCC can further refine the reasoning performance by replacing easy samples with more appropriate hard samples. 
		Such improvements are due to that the DRM addresses a limitation of the CDE and ACE approaches, \ies, the tendency to retain simpler instances while neglecting challenging yet informative samples.

		\begin{figure}[t]
			\centering
			%\vspace{-3em}
			\includegraphics[width=0.475\textwidth]{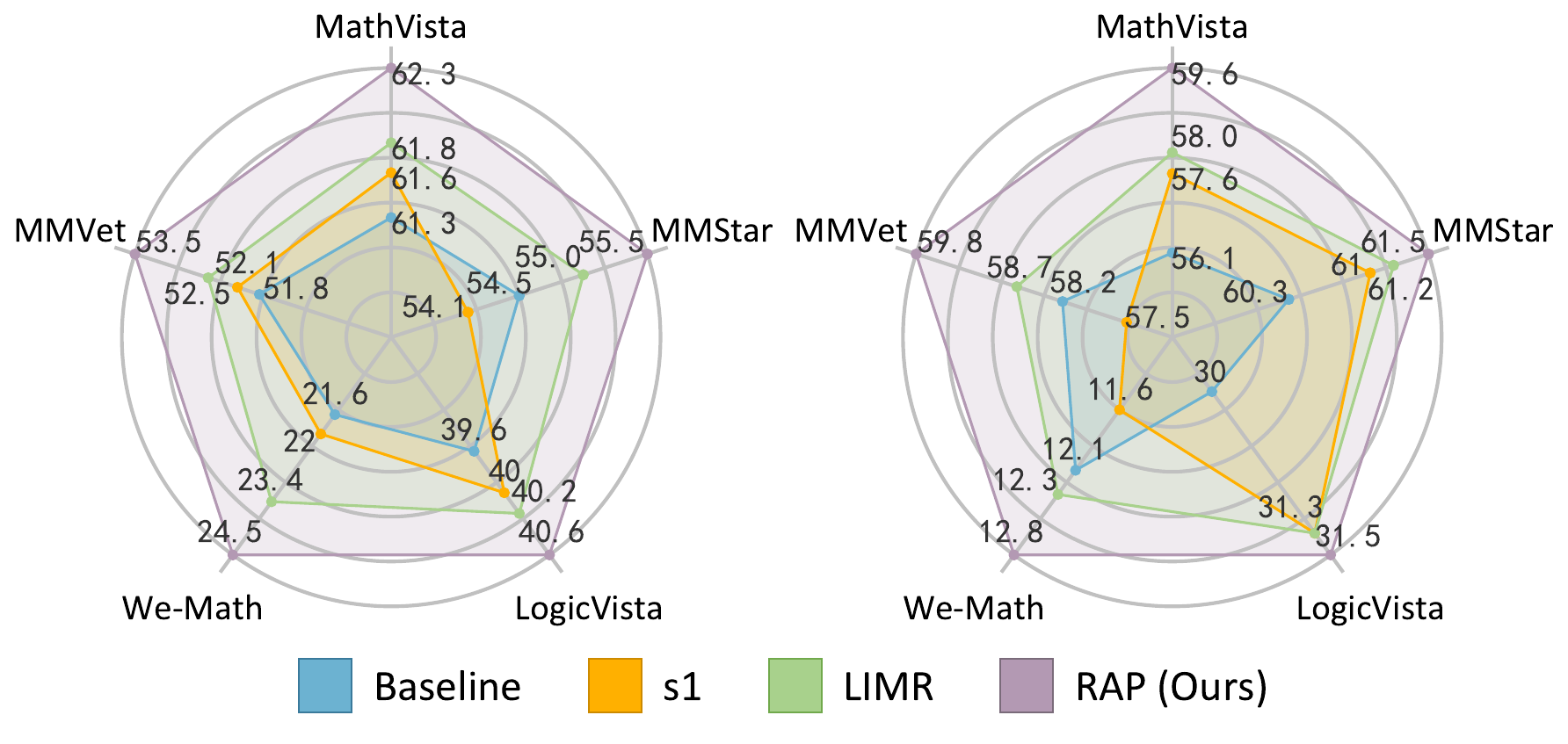}
			\vspace{-3mm}
			\caption{
				Cross-model generalization of cognitive samples selected by RAP. Performance with InternVL3-2B trained on samples from Qwen2.5-VL-3B (left), and vice versa (right).
			}
			\label{fig:leida}
			%\vspace{-3mm}
		\end{figure}
		
		\begin{figure*}[t]
			\centering
			\includegraphics[width=0.99\linewidth]{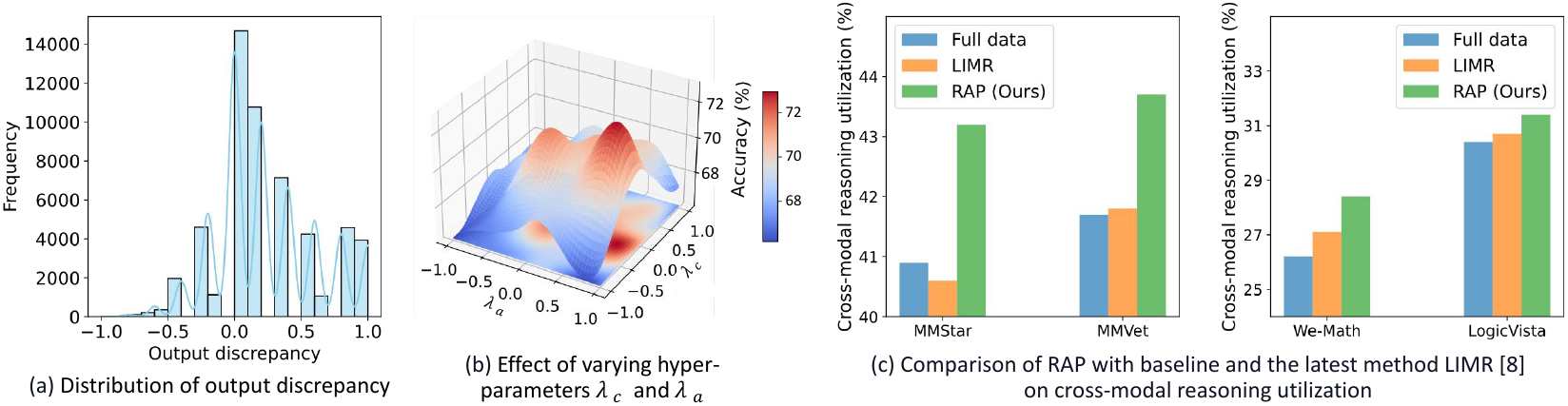}
			\caption{(a) Visualization of output discrepancies between multi-modal and text-only inputs on the full MM-Eureka training dataset. (b) Performance variation with respect to the hyperparameters \( \lambda_{a} \) and \( \lambda_{c} \) on MMstar. (c) Comparative analysis of multi-modal reasoning utilization on four datasets.}
			\label{fig:cross-modal}
			%\vspace{-3mm}
		\end{figure*}

		\noindent\textbf{Comparison on cross-modal reasoning utilization.}
		We evaluate the effectiveness of cognitive samples in enhancing cross-modal reasoning utilization on multi-modal reasoning tasks. 
		We define cross-modal reasoning utilization as the proportion of instances in the MMStar dataset where the model correctly uses multi-modal inputs to answer questions, but fails when relying solely on textual inputs. 
		As shown in Fig. \ref{fig:cross-modal}(c), models trained with cognitive samples show superior integration of cross-modal information, compared to baseline and the latest method LIMR \cite{limr}.    
		Such results highlight that the proposed CDE can effectively reduce models' reliance on superficial linguistic prior by filtering out training samples exhibiting excessive language bias. 
		Hence, models are encouraged to discover the relationship between image and text, thus improving the cross-modal reasoning utilization.

		% \begin{wrapfigure}{r}{0.45\textwidth}
			%     \vspace{-1.5em}
			%     \begin{center}
				%         \includegraphics[width=0.45\textwidth]{image.png}
				%     \end{center}
			%     \vspace{-3mm}
			%     \caption{
				%         Observational data. 
				%         Top: data density of treatment (right) and control (left) groups. 
				%         Middle: observed outcome
				%     }
			%     \label{fig:observational_data}
			%     \vspace{-3mm}
			% \end{wrapfigure}

		\noindent \textbf{Comparison on cross-model generalization.}
		To examine the cross-model generalizability of cognitive samples identified by RAP, we evaluate whether cognitive samples obtained using the Qwen2.5-VL-3b are useful for improving the reasoning of a distinctly structured model, InternVL3-2b, and vice versa.
		Comparative results presented in Fig. \ref{fig:leida} demonstrate that cognitive samples selected by our RAP method outperform the latest method LIMR \cite{limr}, confirming the generalization of RAP in enhancing multi-modal reasoning for varying model frameworks.
		Note that we exclude comparisons with LIMO \cite{limo} due to its similarity with s1 \cite{s1}.

		\noindent\textbf{Analysis on hyperparameter sensitivity.}
		We further investigate the sensitivity of the hyperparameters $\lambda_{c}$ and $\lambda_{a}$ employed within the CDE and ACE components, respectively. 
		As shown in Fig. \ref{fig:cross-modal}(a), we visualize the distribution of output discrepancies between multi-modal and text-only inputs on the full MM-Eureka dataset, revealing a significant presence of samples where multi-modal reasoning is not necessary to solve the task.
		Experimental results depicted in Fig. \ref{fig:cross-modal}(b) suggest that optimal performance is achieved with $\lambda_{c} = 0.1$ and $\lambda_{a} = 0.5$. 
		Performance degradation is observed when both parameters fall below these optimal thresholds, with values lower than 0.1 for $\lambda_{c}$ and 0.5 for $\lambda_{a}$ leading to significant deterioration. 
		This decline is attributed to the inherently uneven distribution of the output-level discrepancy and attention confidence, which are concentrated at the extremes. 
		% As a result, excessively low threshold values hinder the model’s ability to effectively discriminate and filter out biased samples during the selection process.
		
		\noindent\textbf{Analysis on the choice of attention layers.}
		\label{sup:att layer}
		To clarify our rationale for using last-layer attention, we provide additional experiments exploring attention maps from other layers.
		As illustrated in Section \ref{sub:3.2 ace} and Fig. \ref{fig:intro2}(b), attention-biased distributions typically manifest as excessively persistent high attention allocation across subsequent token sequences. 
		Existing research \cite{casualnipsDBLP:conf/nips/RohekarGN23} has demonstrated that attention distributions in the final layer effectively capture causal reasoning behaviors. 
		Such findings intuitively highlight the critical role of the last-layer attention maps in evaluating the reliability of the model's reasoning behavior.  
		Inspired by this, we initially chose the last-layer attention distributions to effectively capture globally biased attention patterns that compromise genuine multi-modal reasoning.
		
		However, we have conducted preliminary experiments to investigate the potential of attention confidence computed from intermediate layers. 
		Recent work \cite{imlayer} has indicated layers 5 to 8 within large language models play a critical role in multi-modal fusion. 
		Therefore, we selected the 6th transformer layer’s attention distribution as an alternative input for our ACE module and compared its efficacy against our original last-layer choice. 
		The results presented in Table \ref{tab:imme layer} demonstrate slight performance degradation when utilizing intermediate (6th-layer) attention maps.
		For most datasets, employing attention maps from the last layer consistently yielded superior performance. 
		These results justify our choice of using last-layer attention maps.
		
		\begin{table}[t]
			\centering
			\caption{Performance comparison between RAP and RAP-mid using QwenVL2.5-3b with the GRPO RL algorithm across three benchmarks. RAP-mid refers to the variant that utilizes intermediate (6th-layer) attention maps.}
			\label{tab:imme layer}
			\footnotesize % Shrink font size to fit space
			\setlength{\tabcolsep}{1.6pt}
			\begin{tabular}{l|ccccc}
				\hline
				Method & MathVista &MMVet &LogicVista &WeMath &Avg. \\
				\hline
				QwenVL2.5-3b-RAP-mid & 64.60 &54.29 &\textbf{41.74} &29.17 &47.45  \\
				\textbf{Qwen2.5VL-3b-RAP (Ours)} & \textbf{64.90} &\textbf{54.63} &41.61 &\textbf{29.33} &\textbf{47.62} \\
				\hline
			\end{tabular}
		\end{table}
		
\begin{figure*}[t]
			\centering
			\includegraphics[width=0.99\linewidth]{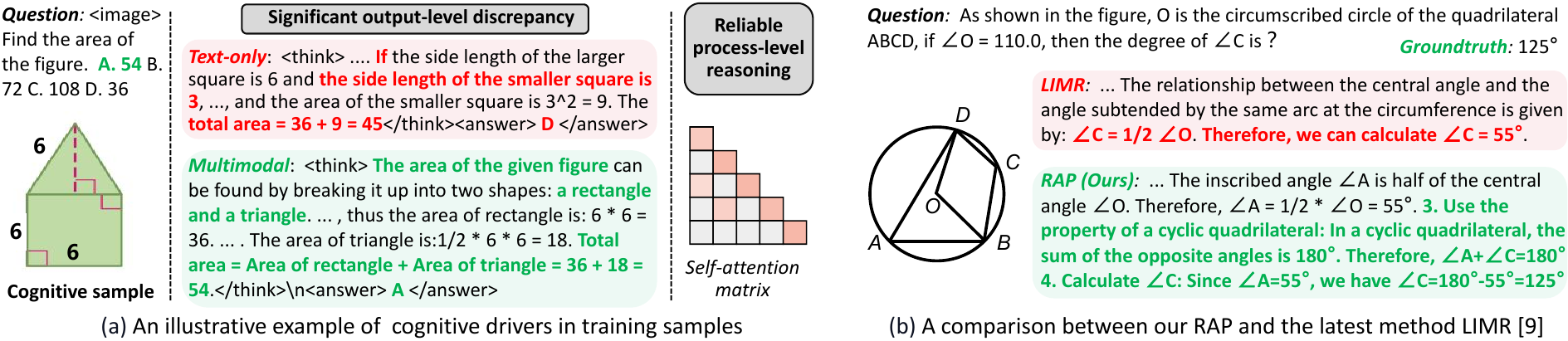}
			\caption{Illustration of (a) characteristics of cognitive samples selected by our RAP method and (b) Comparison of the reasoning processes between our RAP method and the state-of-the-art LIMR \cite{limr}.}
			\label{fig:cog and qua}
			%\vspace{-4mm}
		\end{figure*}

		\noindent\textbf{Qualitative analysis.}
		We provide a qualitative analysis by visualizing cognitive samples and comparing the reasoning processes of our RAP with the latest LIMR method \cite{limr} as follows:
		1) \textit{Visualization of cognitive samples.}
		As shown in Fig. \ref{fig:cog and qua}(a), we present a typical example of the cognitive samples selected by our RAP method, where the model effectively integrates both geometric information and logical reasoning to infer the area.
		This case suggests that the cognitive samples selected by our RAP method exhibit two important characteristics: (a) the necessary of multi-modal information, as evidenced by the significant discrepancy between reasoning outcomes using multi-modal and text-only inputs; and (b) the avoidance of overemphasis on irrelevant or meaningless tokens, ensuring that the model focuses on the most informative features for accurate reasoning.
		2) \textit{Comparison case analysis.}
		As shown in Fig. \ref{fig:cog and qua}(b), we compare the reasoning process of our RAP method with that of the state-of-the-art LIMR approach \cite{limr}. 
		For example, LIMR fails to integrate cross-modal information, leading to incorrect computations of both the central angle and the inscribed angle. 
		In contrast, the model trained using samples selected by RAP correctly applies geometric principles and multi-modal integration to arrive at the correct solution. 
		These comparisons underscore the advantage of training with cognitive samples derived through our RAP method, which enables the model to effectively leverage multi-modal information, resulting in more reliable reasoning outcomes.
		%These show the advantage of cognitive samples selected by RAP, which enables the model to effectively capture multi-modal information.

		\begin{figure*}[t]
			\centering
			\includegraphics[width=0.99\linewidth]{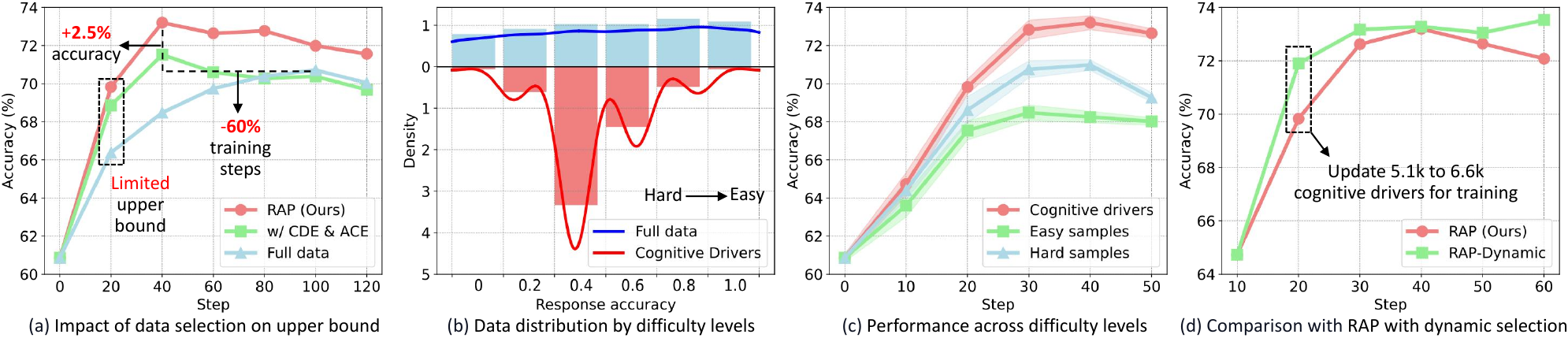}
			\caption{(a) Trade-off analysis between efficiency and performance and effect of data selection on upper bound. (b-c) Impact of varying the proportion of samples with different difficulty levels on reasoning. (c) Comparison with our RAP method and RAP augmented with dynamic selection.}
			\label{fig:dis}
			%\vspace{-5mm}
		\end{figure*}
		
		\subsection{Key Insights and Discussion}
		
		\noindent\textbf{Effect of RAP on reasoning upper bound.}
		As illustrated in Fig. \ref{fig:dis}(a), our RAP method converges faster than the baseline, achieving optimal performance within 40 training steps compared to 100 for the baseline. 
		These results demonstrate RAP's efficiency in enhancing multi-modal reasoning while reducing training overhead and data redundancy. 
		However, further analysis reveals differences between the full RAP model and its variant using only ACE and CDE, particularly during later training stages. 
		This occurs because ACE and CDE inevitably retain easy examples, while discarding challenging yet informative samples.
		For example, multi-modal examples predicted correctly in only one out of five attempts but consistently mispredicted under text-only conditions may be mistakenly eliminated when the discrepancy threshold $\lambda_c$ exceeds 0.2. 
		Such filtering reduces dataset complexity, limiting the model’s overall reasoning upper bound.
		To address this, we propose the Difficulty-aware Replacement Module (DRM) to explicitly replace easy samples with these informative yet challenging examples, thus elevating reasoning performance in later training stages.

		\noindent\textbf{Why does less data outperform the full dataset?}
		The above empirical analysis shows that models trained on carefully curated data can notably outperform those trained on the entire dataset. 
		To further elucidate the underlying mechanisms driving this ``\textit{truth in the few}'' phenomenon, we examine sample difficulty distributions quantified by group-wise response accuracy in the GRPO generation paradigm in Fig. \ref{fig:dis}(b). 
		The results reveal that cognitive samples contain significantly fewer easy samples compared to the full dataset. 
		Therefore, under conventional large-scale training, which usually restricts training to 1 or 2 epochs, the models lack sufficient repeated exposure to challenging samples, thus limiting the improvements brought from these valuable instances. 
		
		Moreover, to further validate our hypothesis, we conduct a comparison in Fig. \ref{fig:dis}(c), where models are trained on equivalent numbers of easy, hard, and cognitive samples. 
		The results reveal that the abundance of easy examples contributes little to advancing the model’s reasoning capability and predominantly introduces redundant information. 
		While challenging samples yield better performance than easy samples, they still fall short compared to the cognitive samples selected through our RAP method. 
		We argue that the superiority of cognitive samples arises not merely from sample difficulty but from their ability to activate the model's multi-modal reasoning capacity. 
		In multi-modal contexts, the training samples must facilitate the activation of the model's multi-modal reasoning mechanisms.

		\begin{table}[t]
			\centering
			\caption{Performance comparison on two text-only benchmarks MATH500 \cite{math500} and AMC23 using Qwen2.5-7B \cite{qwen2025qwen25technicalreport} and LLaMA3.1-8b \cite{llama3.1}, with GRPO as the RL algorithm. Random and Full refer to training with randomly sampled data and the complete dataset, respectively.}
			\label{tab:text-only}
			\small % Shrink font size to fit space
			\setlength{\tabcolsep}{2pt}
			\begin{tabular}{l|cc}
				\hline
				Method & MATH500 & AMC23 \\
				\hline
				Qwen2.5-7b-Random & 72.40 & 50.25  \\
				Qwen2.5-7b-Full   & 74.40 & 51.25   \\
				\textbf{Qwen2.5-7b-RAP (Ours)} & \textbf{74.60} & \textbf{51.75} \\
				\hline
				LLaMA3.1-8b-Random & 46.60 & 20.50  \\
				LLaMA3.1-8b-Full   & 48.60 & 22.75   \\
				\textbf{LLaMA3.1-8b-RAP (Ours)} & \textbf{49.90} & \textbf{23.25} \\
				\hline
			\end{tabular}
		\end{table}
		
		\noindent\textbf{Generalization on text-only scenarios.}
		Our RAP method is primarily designed to enhance multi-modal reasoning during RL post-training, and thus its core CDE relies on cross-modal signals, which is not directly transferable to unimodal settings.
		To address potential concerns regarding applicability to LLMs, we adapt the ACE and DRM modules to text-only scenarios and evaluate them on two text-only benchmarks, MATH500 \cite{math500} and AMC23, using the GRPO \cite{guo2025deepseek} as the RL algorithm.
		We conduct experiments on two representative LLMs, Qwen2.5-7B \cite{qwen2025qwen25technicalreport} and LLaMA3.1-8B \cite{llama3.1}.
		In this adaptation, due to the absence of CDE, we implement DRM by directly removing easy samples defined by $D_\text{iff}=0$, \ie, samples consistently answered correctly across all trials, which typically provide limited optimization signal.
		As shown in Table~\ref{tab:text-only}, ACE and DRM still yield consistent gains over Random and Full training across both models, indicating that these components capture a generally beneficial data selection principle beyond multi-modal settings.
		However, the improvements are smaller than those in the multi-modal scenario.
		We attribute this mainly to two reasons: (1) RAP is originally motivated and optimized for multi-modal contexts where cross-modal discrepancies provide stronger supervision; (2) without CDE, DRM reduces to a difficulty-based filtering strategy, which is less targeted than the original design, thereby limiting the achievable gains.

		\noindent\textbf{Discussion on potential optimization during RL training.}
		The static nature of our current RAP method motivates us to explore potential optimizations through the dynamic adjustment of the training dataset during RL training. 
		Specifically, we examine the evolving distribution of cognitive samples identified by RAP criteria across successive epochs. 
		Initially, as depicted in Fig. \ref{fig:dis}(d), approximately 5,000 samples meet the cognitive sample criteria. 
		However, the number of samples that continues to satisfy this criterion progressively declines as training advances.
		This observed trend suggests the promise of adaptive data sampling, wherein cognitive samples are re-sampled dynamically from the entire dataset after the first and third training epochs. 
		Preliminary experiments indicate that such adaptive strategies yield tangible improvements in model performance. 
		However, despite these advantages, the re-screening of the entire dataset after each epoch incurs a significant computational overhead. 
		As a result, our final approach strategically refrains from adopting this adaptive data selection in order to maintain computational efficiency. 
		This highlights the inherent trade-off between performance enhancements and computational cost when incorporating dynamic data selection strategies into the RL training process.
		
		\section{Conclusion}
		In this work, we introduce a Reasoning Activation Potential (RAP) data selection paradigm, which reduces training costs and improves multi-modal reasoning in MLLMs. 
		RAP utilizes the Causal Discrepancy Estimator (CDE) and Attention Confidence Estimator (ACE) to effectively eliminate attention-biased samples and language-prior biased samples, leading to more accurate and efficient reasoning.
		For future work, we plan to investigate the efficacy of RAP in SFT training and introduce dynamic mechanisms to further improve the efficiency of multi-modal reasoning.

\ifCLASSOPTIONcaptionsoff
  \newpage
\fi

% can use a bibliography generated by BibTeX as a .bbl file
% BibTeX documentation can be easily obtained at:
% http://mirror.ctan.org/biblio/bibtex/contrib/doc/
% The IEEEtran BibTeX style support page is at:
% http://www.michaelshell.org/tex/ieeetran/bibtex/
% \balance{
\small
\bibliographystyle{IEEEtran}
% argument is your BibTeX string definitions and bibliography database(s)
\bibliography{main}

% Generated by IEEEtran.bst, version: 1.14 (2015/08/26)
\begin{thebibliography}{10}
\providecommand{\url}[1]{#1}
\csname url@samestyle\endcsname
\providecommand{\newblock}{\relax}
\providecommand{\bibinfo}[2]{#2}
\providecommand{\BIBentrySTDinterwordspacing}{\spaceskip=0pt\relax}
\providecommand{\BIBentryALTinterwordstretchfactor}{4}
\providecommand{\BIBentryALTinterwordspacing}{\spaceskip=\fontdimen2\font plus
\BIBentryALTinterwordstretchfactor\fontdimen3\font minus
  \fontdimen4\font\relax}
\providecommand{\BIBforeignlanguage}[2]{{%
\expandafter\ifx\csname l@#1\endcsname\relax
\typeout{** WARNING: IEEEtran.bst: No hyphenation pattern has been}%
\typeout{** loaded for the language `#1'. Using the pattern for}%
\typeout{** the default language instead.}%
\else
\language=\csname l@#1\endcsname
\fi
#2}}
\providecommand{\BIBdecl}{\relax}
\BIBdecl

\bibitem{qwen2025qwen25technicalreport}
\BIBentryALTinterwordspacing
Qwen, :, A.~Yang, B.~Yang, B.~Zhang, B.~Hui, B.~Zheng, B.~Yu, C.~Li, D.~Liu,
  F.~Huang, H.~Wei, H.~Lin, J.~Yang, J.~Tu, J.~Zhang, J.~Yang, J.~Yang,
  J.~Zhou, J.~Lin, K.~Dang, K.~Lu, K.~Bao, K.~Yang, L.~Yu, M.~Li, M.~Xue,
  P.~Zhang, Q.~Zhu, R.~Men, R.~Lin, T.~Li, T.~Tang, T.~Xia, X.~Ren, X.~Ren,
  Y.~Fan, Y.~Su, Y.~Zhang, Y.~Wan, Y.~Liu, Z.~Cui, Z.~Zhang, and Z.~Qiu,
  ``Qwen2.5 technical report,'' 2025. [Online]. Available:
  \url{https://arxiv.org/abs/2412.15115}
\BIBentrySTDinterwordspacing

\bibitem{openai2023gpt4}
OpenAI, ``Gpt-4 technical report,'' 2023.

\bibitem{guo2025deepseek}
D.~Guo, D.~Yang, H.~Zhang, J.~Song, R.~Zhang, R.~Xu, Q.~Zhu, S.~Ma, P.~Wang,
  X.~Bi \emph{et~al.}, ``Deepseek-r1: Incentivizing reasoning capability in
  llms via reinforcement learning,'' \emph{arXiv preprint arXiv:2501.12948},
  2025.

\bibitem{hu2024openrlhf}
J.~Hu, X.~Wu, Z.~Zhu, Xianyu, W.~Wang, D.~Zhang, and Y.~Cao, ``Openrlhf: An
  easy-to-use, scalable and high-performance rlhf framework,'' \emph{arXiv
  preprint arXiv:2405.11143}, 2024.

\bibitem{r1vDBLP:journals/corr/abs-2503-10615}
Y.~Yang, X.~He, H.~Pan, X.~Jiang, Y.~Deng, X.~Yang, H.~Lu, D.~Yin, F.~Rao,
  M.~Zhu, B.~Zhang, and W.~Chen, ``R1-onevision: Advancing generalized
  multimodal reasoning through cross-modal formalization,'' \emph{CoRR}, vol.
  abs/2503.10615, 2025.

\bibitem{openai2024reasoning}
\BIBentryALTinterwordspacing
OpenAI, ``Learning to reason with llms,'' 2024. [Online]. Available:
  \url{https://openai.com/index/learning-to-reason-with-llms/}
\BIBentrySTDinterwordspacing

\bibitem{kimiteam2025kimik15scalingreinforcement}
\BIBentryALTinterwordspacing
K.~Team, A.~Du, B.~Gao, B.~Xing, C.~Jiang, C.~Chen, C.~Li, C.~Xiao, C.~Du,
  C.~Liao, C.~Tang, C.~Wang, D.~Zhang, E.~Yuan, E.~Lu, F.~Tang, F.~Sung,
  G.~Wei, G.~Lai, H.~Guo, H.~Zhu, H.~Ding, H.~Hu, H.~Yang, H.~Zhang, H.~Yao,
  H.~Zhao, H.~Lu, H.~Li, H.~Yu, H.~Gao, H.~Zheng, H.~Yuan, J.~Chen, J.~Guo,
  J.~Su, J.~Wang, J.~Zhao, J.~Zhang, J.~Liu, J.~Yan, J.~Wu, L.~Shi, L.~Ye,
  L.~Yu, M.~Dong, N.~Zhang, N.~Ma, Q.~Pan, Q.~Gong, S.~Liu, S.~Ma, S.~Wei,
  S.~Cao, S.~Huang, T.~Jiang, W.~Gao, W.~Xiong, W.~He, W.~Huang, W.~Wu, W.~He,
  X.~Wei, X.~Jia, X.~Wu, X.~Xu, X.~Zu, X.~Zhou, X.~Pan, Y.~Charles, Y.~Li,
  Y.~Hu, Y.~Liu, Y.~Chen, Y.~Wang, Y.~Liu, Y.~Qin, Y.~Liu, Y.~Yang, Y.~Bao,
  Y.~Du, Y.~Wu, Y.~Wang, Z.~Zhou, Z.~Wang, Z.~Li, Z.~Zhu, Z.~Zhang, Z.~Wang,
  Z.~Yang, Z.~Huang, Z.~Huang, Z.~Xu, and Z.~Yang, ``Kimi k1.5: Scaling
  reinforcement learning with llms,'' 2025. [Online]. Available:
  \url{https://arxiv.org/abs/2501.12599}
\BIBentrySTDinterwordspacing

\bibitem{s1}
N.~Muennighoff, Z.~Yang, W.~Shi, X.~L. Li, L.~Fei{-}Fei, H.~Hajishirzi,
  L.~Zettlemoyer, P.~Liang, E.~J. Cand{\`{e}}s, and T.~Hashimoto, ``s1: Simple
  test-time scaling,'' \emph{CoRR}, vol. abs/2501.19393, 2025.

\bibitem{limr}
X.~Li, H.~Zou, and P.~Liu, ``{LIMR:} less is more for {RL} scaling,''
  \emph{CoRR}, vol. abs/2502.11886, 2025.

\bibitem{lp1}
Y.~Han, L.~Nie, J.~Yin, J.~Wu, and Y.~Yan, ``Visual perturbation-aware
  collaborative learning for overcoming the language prior problem,''
  \emph{CoRR}, vol. abs/2207.11850, 2022.

\bibitem{vcd}
S.~Leng, H.~Zhang, G.~Chen, X.~Li, S.~Lu, C.~Miao, and L.~Bing, ``Mitigating
  object hallucinations in large vision-language models through visual
  contrastive decoding,'' in \emph{CVPR}, 2024, pp. 13\,872--13\,882.

\bibitem{limo}
Y.~Ye, Z.~Huang, Y.~Xiao, E.~Chern, S.~Xia, and P.~Liu, ``{LIMO:} less is more
  for reasoning,'' \emph{CoRR}, vol. abs/2502.03387, 2025.

\bibitem{rlhf1}
M.~G. Azar, Z.~D. Guo, B.~Piot, R.~Munos, M.~Rowland, M.~Valko, and
  D.~Calandriello, ``A general theoretical paradigm to understand learning from
  human preferences,'' in \emph{International Conference on Artificial
  Intelligence and Statistics}, vol. 238.\hskip 1em plus 0.5em minus
  0.4em\relax {PMLR}, 2024, pp. 4447--4455.

\bibitem{tmm1}
B.~Lin, Z.~Tang, Y.~Ye, J.~Huang, J.~Zhang, Y.~Pang, P.~Jin, M.~Ning, J.~Luo,
  and L.~Yuan, ``Moe-llava: Mixture of experts for large vision-language
  models,'' \emph{IEEE Transactions on Multimedia}, pp. 1--14, 2026.

\bibitem{tmm2}
H.~Xiong, Y.~Zhuge, J.~Zhu, L.~Zhang, and H.~Lu, ``3ur-llm: An end-to-end
  multimodal large language model for 3d scene understanding,'' \emph{IEEE
  Transactions on Multimedia}, vol.~27, pp. 2899--2911, 2025.

\bibitem{tmm3}
W.~Tang, Y.~Sun, Q.~Gu, and Z.~Li, ``Visual position prompt for mllm based
  visual grounding,'' \emph{IEEE Transactions on Multimedia}, pp. 1--16, 2026.

\bibitem{step-dpo}
X.~Lai, Z.~Tian, Y.~Chen, S.~Yang, X.~Peng, and J.~Jia, ``Step-dpo: Step-wise
  preference optimization for long-chain reasoning of llms,'' \emph{CoRR}, vol.
  abs/2406.18629, 2024.

\bibitem{PR-DPO}
\BIBentryALTinterwordspacing
S.~R. Chowdhury, A.~Kini, and N.~Natarajan, ``Provably robust {DPO:} aligning
  language models with noisy feedback,'' in \emph{Forty-first International
  Conference on Machine Learning, {ICML} 2024, Vienna, Austria, July 21-27,
  2024}.\hskip 1em plus 0.5em minus 0.4em\relax OpenReview.net, 2024. [Online].
  Available: \url{https://openreview.net/forum?id=yhpDKSw7yA}
\BIBentrySTDinterwordspacing

\bibitem{dpo}
R.~Rafailov, A.~Sharma, E.~Mitchell, C.~D. Manning, S.~Ermon, and C.~Finn,
  ``Direct preference optimization: Your language model is secretly a reward
  model,'' in \emph{NeurIPS}, 2023.

\bibitem{ppo}
J.~Schulman, F.~Wolski, P.~Dhariwal, A.~Radford, and O.~Klimov, ``Proximal
  policy optimization algorithms,'' vol. abs/1707.06347, 2017.

\bibitem{rloo}
A.~Ahmadian, C.~Cremer, M.~Gall{\'{e}}, M.~Fadaee, J.~Kreutzer, O.~Pietquin,
  A.~{\"{U}}st{\"{u}}n, and S.~Hooker, ``Back to basics: Revisiting
  reinforce-style optimization for learning from human feedback in llms,'' in
  \emph{ACL}, 2024, pp. 12\,248--12\,267.

\bibitem{lmmr1}
Y.~Peng, G.~Zhang, M.~Zhang, Z.~You, J.~Liu, Q.~Zhu, K.~Yang, X.~Xu, X.~Geng,
  and X.~Yang, ``{LMM-R1:} empowering 3b lmms with strong reasoning abilities
  through two-stage rule-based {RL},'' \emph{CoRR}, vol. abs/2503.07536, 2025.

\bibitem{vrftliu2025visual}
Z.~Liu, Z.~Sun, Y.~Zang, X.~Dong, Y.~Cao, H.~Duan, D.~Lin, and J.~Wang,
  ``Visual-rft: Visual reinforcement fine-tuning,'' \emph{arXiv preprint
  arXiv:2503.01785}, 2025.

\bibitem{mpowang2024enhancing}
W.~Wang, Z.~Chen, W.~Wang, Y.~Cao, Y.~Liu, Z.~Gao, J.~Zhu, X.~Zhu, L.~Lu,
  Y.~Qiao \emph{et~al.}, ``Enhancing the reasoning ability of multimodal large
  language models via mixed preference optimization,'' \emph{arXiv preprint
  arXiv:2411.10442}, 2024.

\bibitem{mulberry}
H.~Yao, J.~Huang, W.~Wu, J.~Zhang, Y.~Wang, S.~Liu, Y.~Wang, Y.~Song, H.~Feng,
  L.~Shen, and D.~Tao, ``Mulberry: Empowering {MLLM} with o1-like reasoning and
  reflection via collective monte carlo tree search,'' \emph{CoRR}, vol.
  abs/2412.18319, 2024.

\bibitem{MM-Eureka}
F.~Meng, L.~Du, Z.~Liu, Z.~Zhou, Q.~Lu, D.~Fu, B.~Shi, W.~Wang, J.~He,
  K.~Zhang, P.~Luo, Y.~Qiao, Q.~Zhang, and W.~Shao, ``Mm-eureka: Exploring
  visual aha moment with rule-based large-scale reinforcement learning,''
  \emph{CoRR}, vol. abs/2503.07365, 2025.

\bibitem{TMM-lss}
S.~Li, X.~Xu, W.~Meng, J.~Song, C.~Peng, and H.~T. Shen, ``Mitigating
  hallucinations in large vision-language models via reasoning
  uncertainty-guided refinement,'' \emph{{IEEE} Trans. Multim.}, vol.~27, pp.
  7380--7391, 2025.

\bibitem{casualnipsDBLP:conf/nips/RohekarGN23}
R.~Y. Rohekar, Y.~Gurwicz, and S.~Nisimov, ``Causal interpretation of
  self-attention in pre-trained transformers,'' in \emph{NeurIPS}, 2023.

\bibitem{casualrw2DBLP:conf/aaai/ZhangZZ24}
C.~Zhang, L.~Zhang, and D.~Zhou, ``Causal walk: Debiasing multi-hop fact
  verification with front-door adjustment,'' in \emph{AAAI}, 2024, pp.
  19\,533--19\,541.

\bibitem{r4_tmm2025}
S.~Zhou, J.~Xiao, X.~Yang, P.~Song, D.~Guo, A.~Yao, M.~Wang, and T.~Chua,
  ``Scene-text grounding for text-based video question answering,'' \emph{TMM},
  2025.

\bibitem{tim1}
L.~Z. Y. B. G. W. Y. Y. H. T.~S. Zheng~Wang, Xing~Xu, ``Evidence-based
  multi-feature fusion for adversarial robustness,'' \emph{IEEE Trans. Pattern
  Anal. Mach. Intell.}, vol.~47, no.~10, pp. 8923--8937, 2025.

\bibitem{scholkopf2012causal}
B.~Sch{\"o}lkopf, D.~Janzing, J.~Peters, E.~Sgouritsa, K.~Zhang, and J.~Mooij,
  ``On causal and anticausal learning,'' \emph{arXiv preprint arXiv:1206.6471},
  2012.

\bibitem{pom}
D.~B. Rubin, ``Causal inference using potential outcomes: Design, modeling,
  decisions,'' \emph{Journal of the American statistical Association}, vol.
  100, no. 469, pp. 322--331, 2005.

\bibitem{pom-llm}
T.~Du, A.~Kanodia, H.~Brunborg, K.~Vafa, and S.~Athey, ``{LABOR-LLM:}
  language-based occupational representations with large language models,''
  \emph{CoRR}, vol. abs/2406.17972, 2024.

\bibitem{hintgrpo}
Q.~Huang, L.~Chan, J.~Liu, W.~He, H.~Jiang, M.~Song, J.~Chen, C.~Yao, and
  J.~Song, ``Boosting {MLLM} reasoning with text-debiased hint-grpo,''
  \emph{CoRR}, vol. abs/2503.23905, 2025.

\bibitem{rubin1974estimating}
D.~B. Rubin, ``Estimating causal effects of treatments in randomized and
  nonrandomized studies,'' \emph{Journal of educational Psychology}, vol.~66,
  no.~5, p. 688, 1974.

\bibitem{ite1}
F.~D. Johansson, U.~Shalit, and D.~A. Sontag, ``Learning representations for
  counterfactual inference,'' in \emph{ICML}, vol.~48, 2016, pp. 3020--3029.

\bibitem{ite2}
U.~Shalit, F.~D. Johansson, and D.~A. Sontag, ``Estimating individual treatment
  effect: generalization bounds and algorithms,'' in \emph{ICML}, vol.~70,
  2017, pp. 3076--3085.

\bibitem{itedrawpearl2009causal}
J.~Pearl, ``Causal inference in statistics: An overview,'' 2009.

\bibitem{itedraw2hammerton2021causal}
G.~Hammerton and M.~R. Munaf{\`o}, ``Causal inference with observational data:
  the need for triangulation of evidence,'' \emph{Psychological medicine},
  vol.~51, no.~4, pp. 563--578, 2021.

\bibitem{CATEabrevaya2015estimating}
J.~Abrevaya, Y.-C. Hsu, and R.~P. Lieli, ``Estimating conditional average
  treatment effects,'' \emph{Journal of Business \& Economic Statistics},
  vol.~33, no.~4, pp. 485--505, 2015.

\bibitem{opera24}
Q.~Huang, X.~Dong, P.~Zhang, B.~Wang, C.~He, J.~Wang, D.~Lin, W.~Zhang, and
  N.~Yu, ``{OPERA:} alleviating hallucination in multi-modal large language
  models via over-trust penalty and retrospection-allocation,'' in \emph{CVPR},
  2024, pp. 13\,418--13\,427.

\bibitem{anctokDBLP:conf/emnlp/WangLDCZMZS23}
L.~Wang, L.~Li, D.~Dai, D.~Chen, H.~Zhou, F.~Meng, J.~Zhou, and X.~Sun, ``Label
  words are anchors: An information flow perspective for understanding
  in-context learning,'' in \emph{EMNLP}, 2023, pp. 9840--9855.

\bibitem{mathvista}
P.~Lu, H.~Bansal, T.~Xia, J.~Liu, C.~Li, H.~Hajishirzi, H.~Cheng, K.~Chang,
  M.~Galley, and J.~Gao, ``Mathvista: Evaluating mathematical reasoning of
  foundation models in visual contexts,'' in \emph{ICLR}, 2024.

\bibitem{mmstar}
L.~Chen, J.~Li, X.~Dong, P.~Zhang, Y.~Zang, Z.~Chen, H.~Duan, J.~Wang, Y.~Qiao,
  D.~Lin, and F.~Zhao, ``Are we on the right way for evaluating large
  vision-language models?'' in \emph{NeurIPS}, 2024.

\bibitem{mathverse}
R.~Zhang, D.~Jiang, Y.~Zhang, H.~Lin, Z.~Guo, P.~Qiu, A.~Zhou, P.~Lu, K.~Chang,
  Y.~Qiao, P.~Gao, and H.~Li, ``{MATHVERSE:} does your multi-modal {LLM} truly
  see the diagrams in visual math problems?'' in \emph{ECCV}, 2024, pp.
  169--186.

\bibitem{wemath}
R.~Qiao, Q.~Tan, G.~Dong, M.~Wu, C.~Sun, X.~Song, Z.~Gongque, S.~Lei, Z.~Wei,
  M.~Zhang, R.~Qiao, Y.~Zhang, X.~Zong, Y.~Xu, M.~Diao, Z.~Bao, C.~Li, and
  H.~Zhang, ``We-math: Does your large multimodal model achieve human-like
  mathematical reasoning?'' \emph{CoRR}, vol. abs/2407.01284, 2024.

\bibitem{mmvet}
W.~Yu, Z.~Yang, L.~Li, J.~Wang, K.~Lin, Z.~Liu, X.~Wang, and L.~Wang, ``Mm-vet:
  Evaluating large multimodal models for integrated capabilities,'' in
  \emph{ICML}, 2024.

\bibitem{logicvista}
Y.~Xiao, E.~Sun, T.~Liu, and W.~Wang, ``Logicvista: Multimodal {LLM} logical
  reasoning benchmark in visual contexts,'' \emph{CoRR}, vol. abs/2407.04973,
  2024.

\bibitem{easyr1}
G.~Sheng, C.~Zhang, Z.~Ye, X.~Wu, W.~Zhang, R.~Zhang, Y.~Peng, H.~Lin, and
  C.~Wu, ``Hybridflow: A flexible and efficient rlhf framework,'' \emph{arXiv
  preprint arXiv: 2409.19256}, 2024.

\bibitem{kwon2023efficient}
W.~Kwon, Z.~Li, S.~Zhuang, Y.~Sheng, L.~Zheng, C.~H. Yu, J.~E. Gonzalez,
  H.~Zhang, and I.~Stoica, ``Efficient memory management for large language
  model serving with pagedattention,'' in \emph{Proceedings of the ACM SIGOPS
  29th Symposium on Operating Systems Principles}, 2023.

\bibitem{internvl}
Z.~Chen, J.~Wu, W.~Wang, W.~Su, G.~Chen, S.~Xing, M.~Zhong, Q.~Zhang, X.~Zhu,
  L.~Lu, B.~Li, P.~Luo, T.~Lu, Y.~Qiao, and J.~Dai, ``Internvl: Scaling up
  vision foundation models and aligning for generic visual-linguistic tasks,''
  \emph{CVPR}, 2024.

\bibitem{imlayer}
H.~Yin, G.~Si, and Z.~Wang, ``Clearsight: Visual signal enhancement for object
  hallucination mitigation in multimodal large language models,'' in
  \emph{CVPR}, 2025, pp. 14\,625--14\,634.

\bibitem{math500}
D.~Hendrycks, C.~Burns, S.~Kadavath, A.~Arora, S.~Basart, E.~Tang, D.~Song, and
  J.~Steinhardt, ``Measuring mathematical problem solving with the {MATH}
  dataset,'' in \emph{NeurIPS Datasets and Benchmarks}, 2021.

\bibitem{llama3.1}
A.~Dubey, A.~Jauhri, A.~Pandey, A.~Kadian, A.~Al-Dahle, A.~Letman, A.~Mathur,
  A.~Schelten, A.~Yang, A.~Fan \emph{et~al.}, ``The llama 3 herd of models,''
  \emph{arXiv e-prints}, pp. arXiv--2407, 2024.

\end{thebibliography}

\end{document}